\documentclass[letterpaper]{article} 
\usepackage[preprint]{aaai2027}  
\usepackage[hyphens]{url}  
\usepackage{graphicx} 
\usepackage{natbib}  
\usepackage{caption} 
\usepackage{algorithm}
\usepackage{algorithmic}
\usepackage{amsmath,amssymb}
\usepackage{pifont}
\newcommand{\cmark}{\ding{51}} 
\newcommand{\xmark}{\ding{55}} 
\usepackage[table]{xcolor}
\usepackage{makecell}
\usepackage{newfloat}
\usepackage{listings}
\DeclareCaptionStyle{ruled}{labelfont=normalfont,labelsep=colon,strut=off} 
\floatstyle{ruled}
\newfloat{listing}{tb}{lst}{}
\floatname{listing}{Listing}

\usepackage{booktabs}

\newcommand{\method}{JEPA-WAM}

\newcommand{\ours}{\textbf{JEPA-WAM}}

\usepackage{array}
\usepackage{makecell}
\newcommand{\intern}{\textsuperscript{\ensuremath{\ddagger}}}
\definecolor{linkblue}{RGB}{0,76,153}
\title{JEPA-WAM: Learning Vision-Language-Action Policies with Joint-Embedding World Modeling}
\author{
    Yihan Lin\textsuperscript{\rm 1,\rm 3}\equalcontrib\intern,
    Jiawei He\textsuperscript{\rm 2}\equalcontrib,
    Shifeng Bao\textsuperscript{\rm 1,\rm 3}\intern,
    Chen Zhao\textsuperscript{\rm 1,\rm 3},
    Yang Li\textsuperscript{\rm 1,\rm 3},
    Xiaobo Wang\textsuperscript{\rm 5},
    Yan Wang\textsuperscript{\rm 6},
    Cheng Chi\textsuperscript{\rm 1,\rm 4}\corresponding,
    Jing Zhang\textsuperscript{\rm 1,\rm 4}\corresponding
}

\affiliations{
    \textsuperscript{\rm 1}School of Information, Renmin University of China, Beijing, China\\
    \textsuperscript{\rm 2}XYZ Embodied AI, Beijing, China\\
    \textsuperscript{\rm 3}Key Laboratory of Data Engineering and Knowledge Engineering, Beijing, China\\
    \textsuperscript{\rm 4}Engineering Research Center of Database and Business Intelligence, Beijing, China\\
    \textsuperscript{\rm 5}Shenzhen University of Advanced Technology, Shenzhen, China\\
    \textsuperscript{\rm 6}Institute for AI Industry Research (AIR), Tsinghua University, Beijing, China\\
}

\begin{document}

\maketitle

\begingroup
\renewcommand{\thefootnote}{\fnsymbol{footnote}}
\footnotetext[3]{Done as intern at XYZ Embodied AI.}
\endgroup

\begin{abstract}
Robust robot control benefits from explicitly modeling state transitions, but video-generation world action models (WAMs) introduce substantial deployment cost.
Existing latent WAMs avoid explicit future generation, but often compress predictive representations or separate predictive modeling from the representations used for action generation.
We introduce JEPA-WAM, a latent WAM built in a pretrained V-JEPA space, which couples latent transition prediction with continuous action generation through a shared predictor.
JEPA-WAM predicts a spatially structured joint current–future target that captures task-shared visual temporal structure between current and future observations, while preserving dense patch-level correspondence. 
Through the shared predictor, transition supervision directly shapes the backbone, from which dedicated representations are extracted for action prediction.
The same design can also be instantiated in pretrained VLA policies while preserving their original perception and action pathways.
On LIBERO-Plus, JEPA-WAM achieves 79.2\%, the best result without large-scale robot-policy pretraining, while its pretrained $\pi_{0.5}$ instantiation reaches 86.3\%, achieving the best overall performance.
Experiments on RoboTwin 2.0 and real-world bimanual manipulation further demonstrate strong generalization under visual and spatial shifts.
The project page is available on
\pdfstartlink
attr{/Border [0 0 0]}
user{/Subtype /Link /A << /S /URI /URI (https://spritewithoutice.github.io/JEPA_WAM/) >>}
\textcolor{linkblue}{GitHub}%
\pdfendlink.
\end{abstract}


\section{Introduction}


Vision-language-action (VLA) policies have achieved strong performance across diverse manipulation tasks~\cite{kim2024openvla,black2024pi_0,liu2025rdt}, but their action prediction objectives model state transitions only implicitly, which can limit robustness under distribution shift.
World action models (WAMs) address this by explicitly modeling future states alongside action generation~\cite{cen2025worldvla,li2026causal,ye2026world}.
However, video generation based WAMs incur substantial deployment cost due to iterative future prediction.
This motivates latent WAMs, which retain predictive world modeling without generating future observations.

\begin{figure}[t]
    \centering
    \includegraphics[width=\columnwidth]{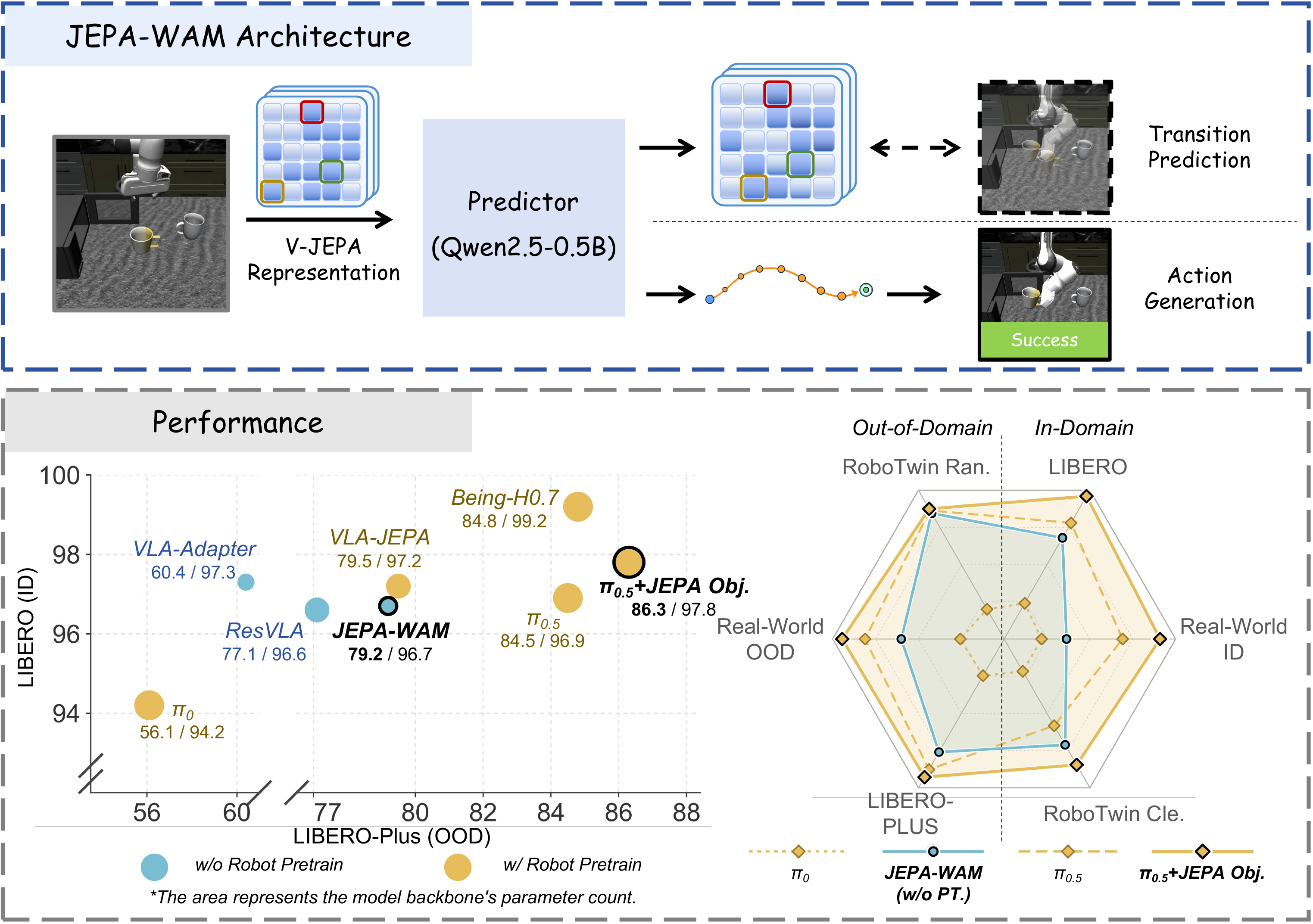}
    \caption{Overview and performance of JEPA-WAM.
Top: JEPA-WAM couples latent transition prediction in the V-JEPA space with action generation through a shared predictor.
Bottom: JEPA-WAM maintains strong in-distribution performance while improving generalization under out-of-distribution shifts across LIBERO, RoboTwin 2.0, and real-world manipulation; the same transition supervision also benefits pretrained $\pi_{0.5}$.
}
    \label{fig:teaser}
\end{figure}

Without explicit future-frame generation, latent WAMs must address two complementary design questions: \textbf{\textit{what predictive target should be learned}}, and \textbf{\textit{how should predictive supervision be integrated with action generation?}} 
Regarding the target, existing methods often reuse intermediate features from pretrained video generators~\citep{yuan2026fastwam} or compress future observations into a small number of latent tokens or subgoals~\citep{luo2026beingh07,chen2026lawam}. 
Although efficient, generator features are optimized for iterative future-frame generation rather than explicitly representing state changes, while compact future representations may lose fine-grained spatial structure.
Regarding policy integration, existing approaches either use predicted future representations as additional context for the action module~\citep{ma2026dit4dit} or introduce a separate prediction objective or latent dynamics module alongside the policy~\citep{sun2026vlajepa}. 
The former may expose the action module to redundant future-state information, whereas the latter may only weakly influence the representations from which actions are generated. 
These limitations motivate a latent WAM that learns a spatially structured representation of the observed transition and uses predictive supervision to directly shape the policy backbone responsible for action conditioning.

To address these limitations, we introduce JEPA-WAM, a latent WAM that retains transition modeling without explicit future generation.
To answer the first question about what predictive representation to learn, we build JEPA-WAM in the pretrained V-JEPA 2.1 representation space and construct a target that represents the transition rather than the absolute future state.
V-JEPA's video pretraining produces temporally consistent representations~\citep{mur2026v}, allowing the jointly encoded current--future observations to capture their temporal relation.
Rather than reconstructing a unique future observation, the joint target captures stable and changing regions and evolving local object and spatial relations, while preserving dense patch-level structure instead of being globally pooled or compressed, allowing the joint target to preserve fine-grained spatial information. 
This formulation can also be applied to pretrained VLA policies; we instantiate it in $\pi_{0.5}$ with an auxiliary transition prediction branch while preserving its original pathways.

To answer the second question about how transition prediction should be used for action generation, JEPA-WAM uses a shared predictor for transition modeling and action generation. 
In a single forward pass, it predicts the dense transition target from the current observation while producing dedicated representations to condition the action generation. 
This allows transition supervision to directly shape the same backbone used for action generation, while the action expert does not need to rely on the full predicted transition representation. 
At deployment, latent transition prediction is removed and only action generation is retained.

As summarized in Figure~\ref{fig:teaser}, JEPA-WAM maintains competitive in-distribution (ID) performance while showing strong generalization under visual and spatial out-of-distribution (OOD) shifts. On LIBERO-Plus, JEPA-WAM achieves 79.2, the best result among methods without robot-policy pretraining. When applied to the transition target in the pretrained $\pi_{0.5}$, it improves the average from 84.5 to 86.3, achieving the best overall result. JEPA-WAM further generalizes well to randomized bimanual manipulation on RoboTwin 2.0 and to real-world manipulation under visual and spatial shifts. Controlled ablations support the joint current--future target, patch-level spatial supervision, and direct transition prediction through the shared backbone.

Our contributions are threefold:
\begin{itemize}

\item We introduce JEPA-WAM, a latent WAM built in a pretrained V-JEPA representation space, where a shared predictor couples latent transition modeling with continuous action generation.

\item We construct a spatially structured joint current--future transition target that preserves dense patch-level information, and instantiate the same target formulation in pretrained VLA policies.

\item We demonstrate strong OOD generalization on LIBERO-Plus, RoboTwin 2.0, and real-world manipulation, with the pretrained $\pi_{0.5}$ instantiation achieving the best overall result on LIBERO-Plus.

\end{itemize}


\section{Related Work}

\begin{figure}[!t]
    \centering
    \includegraphics[width=\columnwidth]{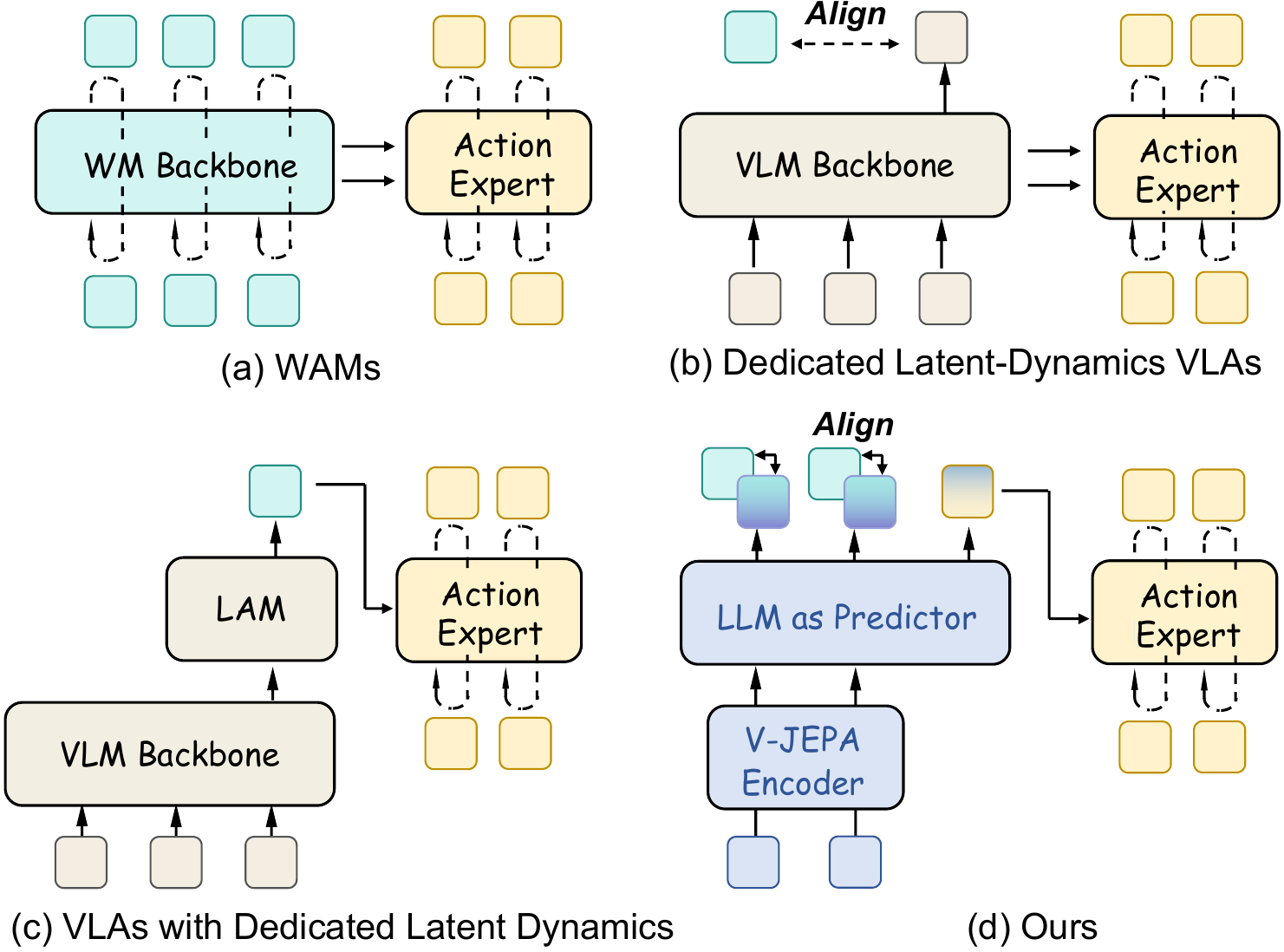}
    \caption{Comparison of latent WAM paradigms. JEPA-WAM couples transition prediction and action generation through a shared predictor.}
    \label{fig:related_work}
\end{figure}

\begin{figure*}[t]
    \centering
    \includegraphics[width=0.95\textwidth]{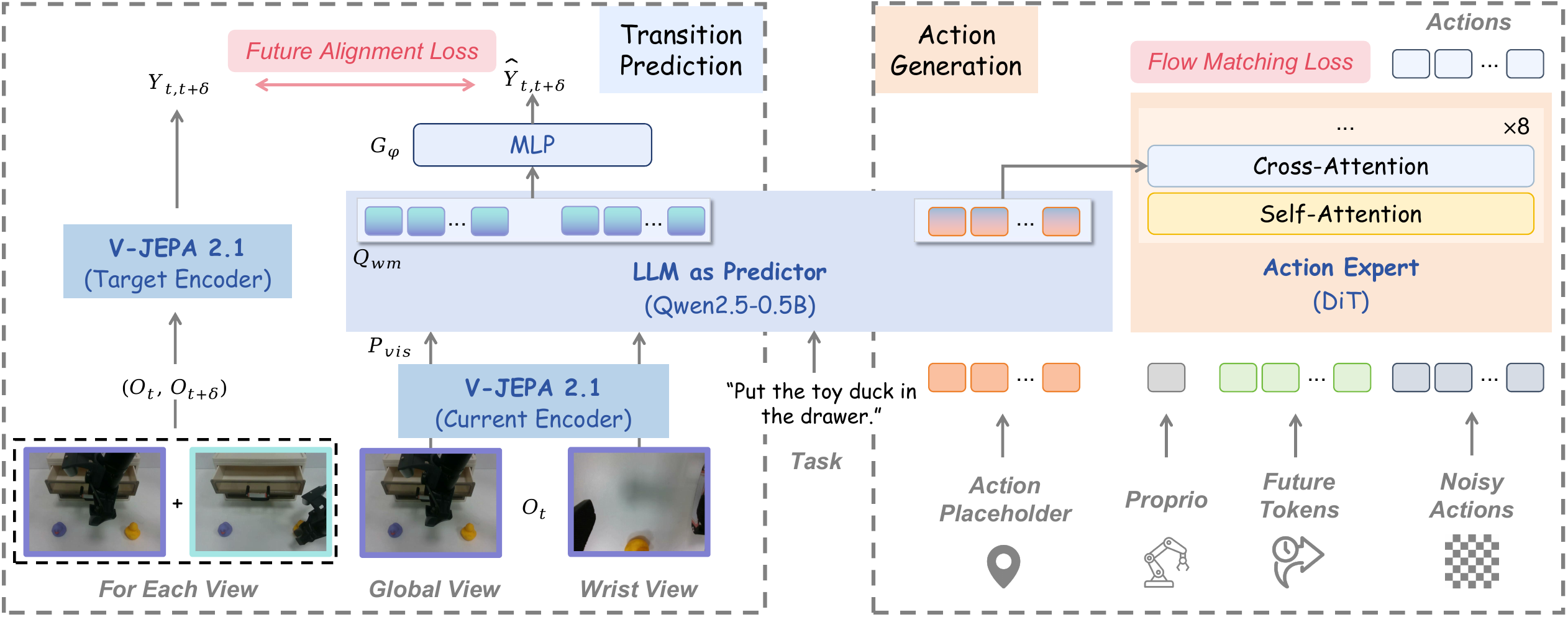}
    \caption{Overview of JEPA-WAM.
A frozen V-JEPA encoder constructs a spatially structured joint current--future target for latent transition prediction.
The shared predictor learns task-shared visual temporal structure from the current observation, while subsequent action representations integrate the visual context and task instruction to condition the action expert.
The target branch is used only during training.}
    \label{fig:framework}
\end{figure*}

\paragraph{Vision-Language-Action Models and World Action Models.}
VLAs combine pretrained vision-language models with action prediction~\citep{intelligence2025pi_,li2026training}. World action models (WAMs) additionally model future states, commonly through future-observation prediction~\citep{cen2025worldvla,li2026causal,ye2026gigaworld}. Recent latent WAMs instead predict future states in representation space without explicit frame generation.

\paragraph{Latent World Action Models.}
Latent WAMs avoid explicit frame generation by moving predictive world modeling into learned representations.
Existing methods broadly follow three directions.
Some reuse intermediate representations from generative world models as context for action prediction~\citep{kim2026cosmospolicy,ye2026gigaworld,li2026lightwam}.
Others compress future observations into latent tokens and train the policy to predict these representations, encouraging the policy to anticipate future states when generating actions~\citep{zheng2025flare,zhao2026frappe,luo2026beingh07}.
A third line uses a dedicated latent dynamics model to predict future representations or latent subgoals, which are then used to condition action generation~\citep{chen2026lawam}.
In contrast, as shown in Figure~\ref{fig:related_work}, JEPA-WAM integrates predictive world modeling and action conditioning through a shared predictor.

\paragraph{JEPA for Robot Policy Learning.}
JEPAs learn predictive representations by matching target embeddings rather than reconstructing pixels~\citep{assran2023ijepa,bardes2024vjepa,assran2025vjepa2,mur2026v}. In robot learning, VLA-JEPA uses a separate JEPA world model to supervise policy representations~\cite{sun2026vlajepa}, while JEPA-VLA provides V-JEPA history representations as additional policy inputs~\cite{miao2026jepa}. Unlike these approaches or action-conditioned JEPA world models for planning, JEPA-WAM builds the latent WAM directly in the V-JEPA representation space: 
V-JEPA represents the current visual state as well as the spatially structured joint current–future transition target, while transition prediction directly supervises the shared action-generating backbone.

\section{Method}
\label{sec:method}

JEPA-WAM is a latent world action model built in a pretrained V-JEPA representation space. 
It uses V-JEPA to represent the current visual state and define the transition target, while a shared predictor couples transition modeling with continuous action generation. 
As shown in Figure~\ref{fig:framework}, it predicts a spatially structured joint current--future target while producing representations for action conditioning.

\subsection{V-JEPA Representation Space and Joint Current--Future Target}
\label{sec:transition}

We use pretrained V-JEPA 2.1 as the latent representation space of JEPA-WAM, representing both the current visual state and the joint current--future target for temporal supervision.
A frozen V-JEPA 2.1 encoder $E_J$ provides dense patch-level representations rather than a globally pooled representation.
We refer to these representations as spatially structured because the tokens retain their patch organization within each view and are arranged in a fixed camera order across views.
We denote the resulting current representation as $Z_t$.
During training, we construct a joint current--future target $Y_{t,t+\delta}$ by encoding the observations at time $t$ and $t+\delta$ together in the same representation space.

\paragraph{Current visual representation.}
Let $V$ index the available camera views, $O_t^v$ denote the current observation from view $v$. The frozen V-JEPA encoder processes each view independently, and we concatenate the resulting visual tokens in a fixed camera order:
\begin{equation}
Z_t =
\operatorname{Concat}_{v\in V} E_J(O_t^v)
\in \mathbb{R}^{N_{\mathrm{vis}}\times d_J},
\end{equation}
where $N_{\mathrm{vis}}$ is the total number of visual tokens and $d_J$ is the V-JEPA feature dimension. 
Thus, V-JEPA defines the latent visual state space of JEPA-WAM, with both the current representation and the transition target below expressed in this space.
The fixed camera and patch-token ordering is retained in the prediction target defined below.

\paragraph{Joint current--future target.}
During training, each current observation is paired with an observation collected $\delta$ steps later. 
For each camera view, we stack the current and future observations along the temporal dimension and jointly encode them with the frozen V-JEPA encoder:
\begin{equation}
\begin{aligned}
Y_{t,t+\delta}
&=
\operatorname{Concat}_{v\in V}
\operatorname{sg}\!\left[
E_J\!\left(
\operatorname{Stack}_{\mathrm{time}}
(O_t^v, O_{t+\delta}^v)
\right)
\right], \\
&\in \mathbb{R}^{N_{\mathrm{vis}}\times d_J}.
\end{aligned}
\end{equation}
where $\operatorname{sg}$ denotes stop-gradient, and $\delta$ is a benchmark-specific temporal offset.

V-JEPA 2.1 uses modality-specific tokenizers, with its video tokenizer grouping every two frames into one temporal tubelet.
Therefore, the two-frame joint input produces the same spatial token grid as a single image, so $Y_{t,t+\delta}$ and $Z_t$ share the same camera and spatial-token ordering.

Unlike a future-only target $E_J(O_{t+\delta}^v)$, which represents the future observation in isolation, the joint target makes both temporal endpoints available to the pretrained V-JEPA encoder. 
Rather than requiring reconstruction of a complete or unique future observation, it emphasizes their visual relation: which regions remain stable or change, and how local object and spatial relations differ across time. 
Together with its dense patch-level organization, this provides task-shared visual temporal supervision without compressing the transition into a small set of global latent tokens.

\subsection{Shared Predictor for Transition Prediction and Action Generation}
\label{sec:shared_predictor}

Having defined the joint current--future target, we next couple its prediction with action generation through a shared predictor.
We instantiate the shared predictor $F_\theta$ with Qwen2.5-0.5B, which processes the visual representation together with the task instruction and produces dedicated representations for action conditioning.
The same predictor is also supervised to predict the joint current--future target, allowing temporal supervision to directly optimize the backbone used for action generation.

\paragraph{Visual interface alignment.}
To bridge the frozen V-JEPA encoder and the Qwen predictor, we introduce a lightweight visual projector $P_{\mathrm{vis}}$ that maps the current V-JEPA representation $Z_t$ into the predictor input space. 
Following the single-stage finetuning setup of Prismatic~\citep{prismaticvlm}, we keep the V-JEPA encoder frozen while jointly finetuning $P_{\mathrm{vis}}$ and the full Qwen2.5-0.5B backbone. During robot-policy training, the V-JEPA encoder, visual projector, and base Qwen weights are frozen, while the Qwen LoRA adapters are optimized together with the transition prediction head and action expert.

\paragraph{Shared predictor.}
To couple latent transition prediction with action generation, we introduce a shared predictor $F_\theta$ that supports both prediction of the joint current--future target and extraction of dedicated representations for action conditioning.
The projected visual tokens, task instruction $\ell$, and dedicated action placeholder tokens $P_{\mathrm{act}}$ are processed by $F_\theta$:
\begin{equation}
(Q_t^{\mathrm{wm}}, C_t)
=
F_\theta(P_{\mathrm{vis}}(Z_t), \ell, P_{\mathrm{act}}),
\end{equation}
where $Q_t^{\mathrm{wm}}$ denotes the hidden states at the visual-token positions and is used to predict the joint current--future target, while $C_t$ denotes the dedicated representations used for action generation. 
$Q_t^{\mathrm{wm}}$ preserves the fixed camera and spatial-token ordering of $Z_t$, whereas $C_t$ aggregates the preceding visual and task context. 
$Q_t^{\mathrm{wm}}$ learns visual temporal structure rather than a complete instruction-conditioned future. 
By sharing the predictor, supervision from latent transition prediction updates the same backbone from which action relevant representations are extracted, allowing the learned temporal visual patterns to benefit action generation.

\paragraph{Latent transition prediction.}
For the transition prediction branch of the shared predictor, the hidden states $Q_t^{\mathrm{wm}}$ produced by $F_\theta$ are mapped back to the V-JEPA representation space through a lightweight prediction head $G_\phi$:
\begin{equation}
\hat{Y}_{t,t+\delta}
=
G_\phi(Q_t^{\mathrm{wm}})
\in
\mathbb{R}^{N_{\mathrm{vis}}\times d_J}.
\end{equation}
Because $Q_t^{\mathrm{wm}}$ preserves the fixed camera and spatial ordering of $Z_t$, $\hat{Y}_{t,t+\delta}$ maintains patch-level correspondence with the joint current--future target $Y_{t,t+\delta}$.
This spatial correspondence allows the model to capture spatially localized changes between the current and future observations at the patch level.
We therefore optimize the mean patch-level cosine distance:
\begin{equation}
\mathcal{L}_{\mathrm{wm}}
=
\frac{1}{B N_{\mathrm{vis}}}
\sum_{b=1}^{B}
\sum_{n=1}^{N_{\mathrm{vis}}}
\left(
1 -
\cos\left(
\hat{Y}_{t,t+\delta,n}^{(b)},
Y_{t,t+\delta,n}^{(b)}
\right)
\right),
\end{equation}
where $B$ denotes the batch size.
Since the V-JEPA encoder and visual projector are frozen and the target representation is stop-gradient, $\mathcal{L}_{\mathrm{wm}}$ updates $F_\theta$ together with the prediction head $G_\phi$, providing patch-level temporal supervision to the shared predictor.

\paragraph{Action prediction and generation.}
The dedicated action placeholders produce $C_t$, which is provided to the DiT action expert $A_\psi$.
Following the StarVLA action-head design~\citep{starvla}, we use conditional flow matching with future tokens and proprioceptive state $s_t$.

Let $a \equiv a_{t:t+H-1}$ denote a demonstrated action chunk.
Given $\epsilon\sim\mathcal{N}(0,I)$ and a flow time $\tau$ sampled using a Beta-based schedule, we define
$a_\tau=(1-\tau)\epsilon+\tau a$ and optimize
\begin{equation}
\mathcal{L}_{\mathrm{act}}
=
\mathbb{E}_{\epsilon,\,\tau}
\left[
\left\|
A_\psi(a_\tau,\tau,s_t,C_t)
-(a-\epsilon)
\right\|_2^2
\right].
\end{equation}

Unless otherwise specified, we use the velocity-prediction objective above. For RoboTwin 2.0, we instead use x-prediction and directly predict the clean action trajectory from the noisy trajectory; see Appendix~\ref{appendix_b_2}.

\subsection{Joint Training and Deployment}
\label{sec:joint_training}

During policy training, we jointly optimize latent transition prediction and action generation:
\begin{equation}
\mathcal{L}
=
\mathcal{L}_{\mathrm{act}}
+
\lambda_{\mathrm{wm}}\mathcal{L}_{\mathrm{wm}},
\end{equation}

\noindent where $\lambda_{\mathrm{wm}}$ balances the two objectives. 
Both losses update the shared predictor $F_\theta$. Detailed optimization settings are provided in Appendix~A.

At deployment, the target branch and prediction head are removed.

\subsection{Transfer to Pretrained VLA Policies}
\label{sec:transfer}

\begin{figure}[t]
    \centering
    \includegraphics[width=\columnwidth]{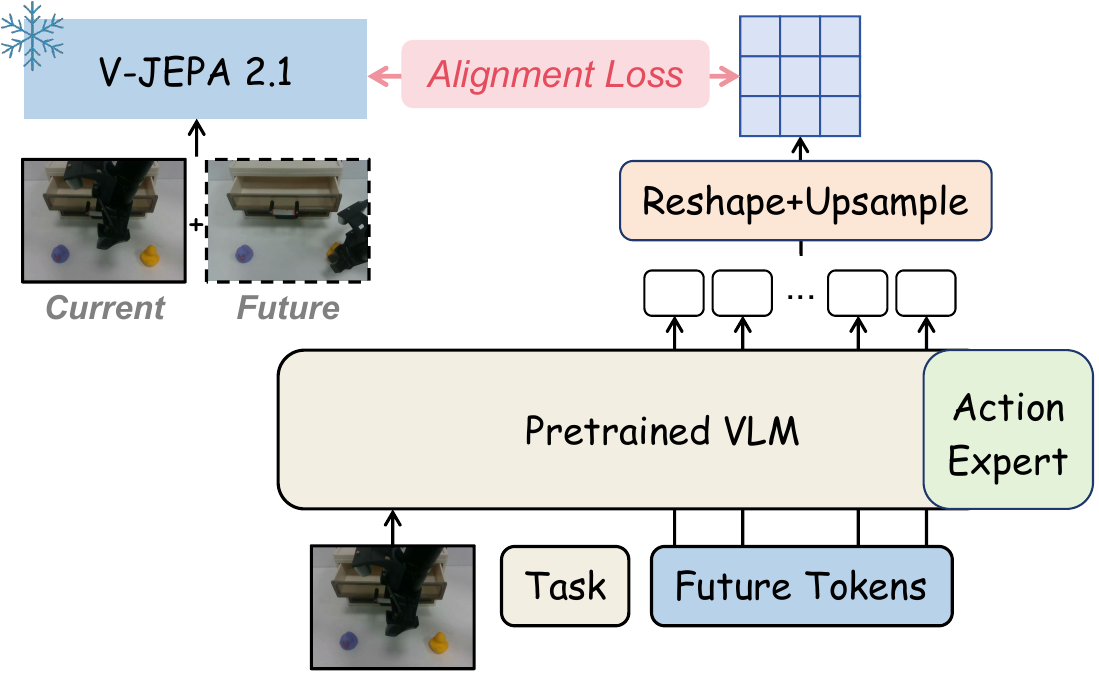}
    \caption{Transfer of the proposed transition supervision to a pretrained VLA policy.}
    \label{fig:transfer}
\end{figure}

The same joint current–future target can supervise pretrained VLAs without modifying their original perception or action pathways (Figure.~\ref{fig:transfer}).

Given a pretrained VLA policy, we introduce a set of future tokens and use their output hidden states to predict the joint current--future target.
Let
$R_t\in\mathbb{R}^{H_fW_f\times d_f}$
denote these hidden states, where $H_f\times W_f$ denotes their coarse spatial grid and $d_f$ is the hidden dimension.
We arrange them as a coarse two-dimensional feature map,
\begin{equation}
\tilde{R}_t
=
\operatorname{Reshape}(R_t)
\in
\mathbb{R}^{H_f\times W_f\times d_f}.
\end{equation}

We first reshape the future-token representations into a coarse two-dimensional feature map, then apply a lightweight projection and spatial upsampling to match the feature dimension and spatial resolution of the joint current--future target $Y_{t,t+\delta}$:
\begin{equation}
\hat{Y}_{t,t+\delta}
=
\operatorname{Upsample}
\left(
P_{\mathrm{sp}}(\tilde{R}_t)
\right)
\in
\mathbb{R}^{N_{\mathrm{vis}}\times d_J}.
\end{equation}
where $P_{\mathrm{sp}}$ denotes the lightweight projection used to match the target feature dimension. This alignment enables the pretrained policy to receive the same patch-level supervision from the joint current--future target while preserving its original action pathways. Additional implementation details are provided in Appendix~A.4.

\section{Experiments}
\label{sec:experiments}

We organize the experiments around three questions:
\smallskip

\textit{\textbf{Question 1.} Does JEPA-WAM improve OOD generalization while preserving competitive ID performance?}
(\S\ref{sec:sim-generalization})

\smallskip

\textit{\textbf{Question 2.} Which design choices are responsible for the observed generalization gains?}
(\S\ref{sec:ablation})

\smallskip

\textit{\textbf{Question 3.} Do these generalization benefits extend to real-world manipulation?}
(\S\ref{sec:real-world})

\smallskip

We first summarize the experimental setup, then evaluate the method on simulation benchmarks, analyze the key designs, and finally examine on real-world manipulation.

\subsection{Experimental Setup}

\paragraph{Benchmarks.}
We evaluate on LIBERO~\citep{liu2023libero}, LIBERO-Plus~\citep{fei2025liberoplus}, RoboTwin 2.0~\citep{chen2025robotwin}, and a real-world AgileX COBOT Magic bimanual platform.
We jointly train on the four LIBERO suites and directly evaluate the same policy on LIBERO-Plus without OOD fine-tuning.
On RoboTwin 2.0, we train on Clean demonstrations from 20 tasks and evaluate the same 20 tasks under both Clean and Random settings.

\paragraph{Training and evaluation.}
JEPA-WAM uses a frozen 300M-parameter V-JEPA 2.1 encoder, a Qwen2.5-0.5B predictor, and a DiT-based flow-matching action expert.
We follow the standard evaluation protocol of each benchmark and report task success rate (\%).
Additional benchmark and evaluation details are provided in Appendix~B.

\subsection{Generalization in Simulation}
\label{sec:sim-generalization}

\paragraph{Standard LIBERO.}
We first evaluate in-distribution performance on LIBERO.
As shown in Table~\ref{tab:libero-standard}, JEPA-WAM achieves an average success rate of 96.7\%, remaining competitive with strong baselines with or without robot-policy pretraining.
Instantiating the same transition supervision in the pretrained $\pi_{0.5}$ further improves its average success rate from 96.9\% to 97.8\%.
These results show that transition modeling preserves strong ID performance for JEPA-WAM and can further benefit a pretrained VLA policy.

\begin{table}[t]
\centering
\small
\setlength{\tabcolsep}{1mm}
\begin{tabular}{@{}
>{\raggedright\arraybackslash}m{26mm}
>{\centering\arraybackslash}m{10mm}
*{5}{>{\centering\arraybackslash}m{6.8mm}}
@{}}
\toprule
Method & Params. & Spa. & Obj. & Goal & Long & Avg. \\
\midrule

\multicolumn{7}{@{}l}{\textit{Without robot-policy pretraining}} \\
\addlinespace[1pt]

Diffusion Policy\\[-1pt]
\citep{chi2024diffusionpolicyvisuomotorpolicy}
& -- & 78.3 & 92.5 & 68.3 & 50.5 & 72.4 \\

ResVLA\\[-1pt]
\citep{zhong2026noiseintentanchoringgenerative}
& 2 & 96.0 & 100.0 & 97.4 & 92.8 & 96.6 \\

Fast-WAM\\[-1pt]
\citep{yuan2026fastwam}
& 5 & 98.2 & 100.0 & 97.0 & 95.2 & \textbf{97.6} \\

\addlinespace[1pt]
\rowcolor{blue!8}
\textbf{\ours{} (Ours)}
& 0.5 & 95.6 & 99.4 & 97.2 & 94.6 & \underline{96.7} \\

\midrule

\multicolumn{7}{@{}l}{\textit{With robot-policy pretraining}} \\
\addlinespace[1pt]

$\pi_{0.5}$\\[-1pt]
\citep{intelligence2025pi_}
& 3 & 98.6 & 98.2 & 98.4 & 92.4 & 96.9 \\

VLA-JEPA\\[-1pt]
\citep{sun2026vlajepa}
& 2 & 96.2 & 99.6 & 97.2 & 95.8 & 97.2 \\

Motus\\[-1pt]
\citep{bi2025motus}
& 5 & 96.8 & 99.8 & 96.6 & 97.6 & \underline{97.7} \\

\addlinespace[1pt]
\rowcolor{blue!8}
\textbf{$\pi_{0.5}$+JEPA Obj. (Ours)}
& 3 & 99.0 & 98.0 & 97.6 & 96.4 & \textbf{97.8} \\

\bottomrule
\end{tabular}
\caption{Success rates (\%) on LIBERO, grouped by the use of large-scale robot-policy pretraining. Params. (B) denotes the scale of the main backbone, excluding the action expert. Best and second-best averages are marked within each group.}
\label{tab:libero-standard}
\end{table}

\begin{table*}[t]
\centering
\small
\setlength{\tabcolsep}{2pt}
\begin{tabular}{@{}lcccccccccc@{}}
\toprule
Method & Params. & PT & Camera & Robot & Language & Light & Back. & Noise & Layout & Avg. \\
\midrule
\multicolumn{11}{@{}l}{\textit{Without robot-policy pretraining}} \\
\addlinespace[1pt]
VLA-Adapter \citep{wang2025vlaadaptereffectiveparadigmtinyscale} & 0.5 & \xmark & 36.2 & 37.9 & 74.6 & 70.6 & 76.1 & 58.0 & 69.7 & 60.4 \\
RoVLA \citep{luo2026rovlamulticonsistencyconstraintsrobust} & 2 & \xmark & \underline{58.4} & 36.3 & \textbf{92.9} & \textbf{95.6} & \textbf{95.0} & \underline{80.9} & 73.0 & 76.0 \\
ResVLA \citep{zhong2026noiseintentanchoringgenerative} & 2 & \xmark & 49.8 & \textbf{59.9} & \underline{88.5} & 90.5 & \underline{94.9} & 76.8 & 79.0 & \underline{77.1} \\
\addlinespace[2pt]
\rowcolor{blue!8}
\textbf{\ours{} (Ours)} & 0.5 & \xmark & \textbf{79.2} & \underline{59.2} & 68.2 & \underline{93.3} & 94.6 & \textbf{83.6} & \underline{76.1} & \textbf{79.2} \\
\midrule
\multicolumn{11}{@{}l}{\textit{With robot-policy pretraining}} \\
\addlinespace[1pt]

VLA-JEPA \citep{sun2026vlajepa}
& 2 & \cmark
& 63.3 & 67.1 & 85.4 & 95.6 & 93.6 & 66.3 & 85.1 & 79.5 \\

PokeVLA \citep{zheng2026pokevla}
& 0.5 & \cmark
& \textbf{84.7} & 46.1 & 84.8 & 94.6 & 82.6 & 89.8 & 77.2 & 80.0 \\

ABot-M0 \citep{yang2026abotm0}
& 4 & \cmark
& 60.4 & 67.9 & \underline{86.4} & 96.2 & 91.6 & 86.4 & 82.6 & 81.6 \\

Cosmos-Policy \citep{kim2026cosmospolicy}
& 2 & \cmark
& 75.8 & 63.3 & 81.7 & 96.5 & 88.9 & \underline{92.7} & 82.2 & 83.0 \\

$\pi_{0.5}$ \citep{intelligence2025pi_}
& 3 & \cmark
& 69.4 & \underline{75.3} & 82.6 & 96.7 & \textbf{96.8} & 84.3 & 86.2 & 84.5 \\

Being-H0.7 \citep{luo2026beingh07}
& 3 & \cmark
& \underline{82.0} & 59.0 & 82.8 & \textbf{97.8} & 90.0 & \textbf{93.5} & \textbf{88.5} & \underline{84.8} \\

\addlinespace[2pt]
\rowcolor{blue!8}
\textbf{$\pi_{0.5}$+JEPA Obj. (Ours)}
& 3 & \cmark
& 66.0 & \textbf{82.0} & \textbf{86.5} & \underline{96.8}
& \underline{96.0} & 88.3 & \underline{88.3} & \textbf{86.3} \\
\bottomrule
\end{tabular}
\caption{Success rates (\%) on LIBERO-Plus after training on LIBERO demonstrations, grouped by large-scale robot-policy pretraining (PT).
Params. denotes the main backbone size excluding the action expert.
Best and second-best averages within each group are bold and underlined.}
\label{tab:libero-plus}
\end{table*}

\begin{table*}[t]
\centering
\small
\begin{tabular}{c|cc|cc|cc|cc|cc|c|cc}
\toprule
Method
& \multicolumn{2}{c|}{Adjust Bottle}
& \multicolumn{2}{c|}{\makecell[c]{Dump Bin\\Bigbin}}
& \multicolumn{2}{c|}{Shake Bottle}
& \multicolumn{2}{c|}{Press Stapler}
& \multicolumn{2}{c|}{\makecell[c]{Stack Bowls\\Two}}
& $\cdots$
& \multicolumn{2}{c}{AVG} \\

& Cle. & Ran.
& Cle. & Ran.
& Cle. & Ran.
& Cle. & Ran.
& Cle. & Ran.
& & Cle. & Ran. \\
\midrule
\multicolumn{14}{@{}l}{\textit{Without robot-policy pretraining}} \\
\addlinespace[1pt]
DP
& 97 & 0
& 49 & 0
& 65 & 8
& 6 & 0
& 61 & 0
& $\cdots$
& 48.0 & 1.6 \\

ACT
& 97 & 23
& 68 & 1
& 74 & 10
& 31 & 6
& 82 & 0
& $\cdots$
& 51.8 & 4.0 \\

DP3
& 99 & 3
& 85 & 53
& 98 & 19
& 69 & 3
& 83 & 6
& $\cdots$
& 73.9 & 8.3 \\

\rowcolor{blue!8}
\textbf{JEPA-WAM}
& 99 & 87
& 94 & 63
& 94 & 55
& 88 & 70
& 94 & 76
& $\cdots$
& \textbf{79.9} & \textbf{36.9} \\

\midrule
\multicolumn{14}{@{}l}{\textit{With robot-policy pretraining}} \\
\addlinespace[1pt]
RDT-1B$^\dagger$
& 81 & 75
& 64 & 32
& 74 & 45
& 41 & 24
& 76 & 30
& $\cdots$
& 56.0 & 24.1 \\

$\pi_0^\dagger$
& 90 & 56
& 83 & 24
& 97 & 60
& 62 & 29
& 91 & 41
& $\cdots$
& 62.5 & 23.9 \\

$\pi_{0.5}^\dagger$
& 98 & 26
& 95 & 41
& 99 & 82
& 66 & 22
& 87 & 40
& $\cdots$
& 75.4 & 37.2 \\

\rowcolor{blue!8}
\makecell[c]{\textbf{$\pi_{0.5}$+JEPA.Obj$^\dagger$} \textbf{(Ours)}}
& 100 & 31
& 90 & 58
& 100 & 84
& 86 & 22
& 93 & 45
& $\cdots$
& \textbf{84.6} & \textbf{37.5} \\

\bottomrule
\end{tabular}
\caption{
Clean (Cle.) and Random (Ran.) success rates (\%) on RoboTwin 2.0.
Selected tasks are shown, while AVG is computed over all 20 tasks.
$\dagger$ denotes large-scale robot-policy pretraining.
}
\label{tab:robotwin}
\end{table*}

\paragraph{LIBERO-Plus.}
We next evaluate OOD generalization on LIBERO-Plus without OOD fine-tuning.
As shown in Table~\ref{tab:libero-plus},  JEPA-WAM achieves 79.2\% average success, the best result among methods without large-scale robot-policy pretraining, while remaining competitive with pretrained VLA and WAM methods.
The same transition supervision also improves the pretrained $\pi_{0.5}$ from 84.5\% to 86.3\%, achieving the best overall result.
These results demonstrate strong scene-level generalization across diverse visual shifts, while the improvement on $\pi_{0.5}$ shows that the proposed transition supervision remains beneficial on top of robot-policy pretraining.

\paragraph{RoboTwin 2.0.}
We further evaluate on RoboTwin 2.0 under both Clean and Random settings.
JEPA-WAM achieves 79.9\% on Clean and 36.9\% on Random, substantially outperforming other methods without robot-policy pretraining and reaching Random performance comparable to the pretrained $\pi_{0.5}$.
The same transition supervision further improves the pretrained
$\pi_{0.5}$ from 75.4\% to 84.6\% on Clean, while maintaining comparable performance on Random (37.5\% vs.\ 37.2\%). Complete results for all 20 tasks are provided in Appendix~D.
JEPA-WAM also runs efficiently at 85 ms per inference (11.76 Hz) on RoboTwin, faster than ABot-M0 (125 ms, 7.99 Hz)~\citep{yang2026abotm0}; see Appendix~C.3.

Together, the results across LIBERO-Plus and RoboTwin 2.0 show that JEPA-WAM maintains competitive ID performance while generalizing strongly to unseen scene configurations and layouts, while the transition supervision also benefits the pretrained $\pi_{0.5}$ across these benchmarks.

\subsection{Design Analysis}
\label{sec:ablation}

\begin{table}[t]
\centering
\small
\setlength{\tabcolsep}{1.7pt}
\renewcommand{\arraystretch}{1.25}
\begin{tabular}{@{}lcccccccc@{}}
\toprule
Method
& Cam. & Rob. & Lang. & Lit.
& Back. & Noi. & Lay. & Avg. \\
\midrule

a. DINO+SigLIP
& 60.0 & 61.9 & 74.1 & 88.7
& 88.0 & 64.2 & 75.7 & 73.2 \\

b. V-JEPA only
& 78.7 & 40.9 & 70.9 & 96.7
& 84.1 & 88.3 & 79.3 & 77.0 \\

c. Future only
& 75.1 & 47.1 & 69.6 & 96.0
& 93.4 & 81.5 & 78.4 & 77.3 \\

d. iREPA align.
& 68.9 & 45.5 & 69.2 & 90.9
& 89.1 & 81.5 & 77.7 & 74.7 \\

e. Lower-16 align.
& 77.5 & 41.6 & 75.0 & 95.5
& 86.3 & 82.2 & 77.2 & 76.5 \\

f. Full hidden
& 62.5 & 49.9 & 70.1 & 89.3
& 88.6 & 75.6 & 76.0 & 73.1 \\

\midrule
\rowcolor{blue!8}
\method{}
& 79.2 & 59.2 & 68.2 & 93.3
& 94.6 & 83.6 & 76.1 & \textbf{79.2} \\
\bottomrule
\end{tabular}
\caption{Category-wise ablation on LIBERO-Plus.}
\label{tab:libero-plus-ablation}
\end{table}

\begin{figure*}[t]
    \centering
    \includegraphics[width=1.0\textwidth]{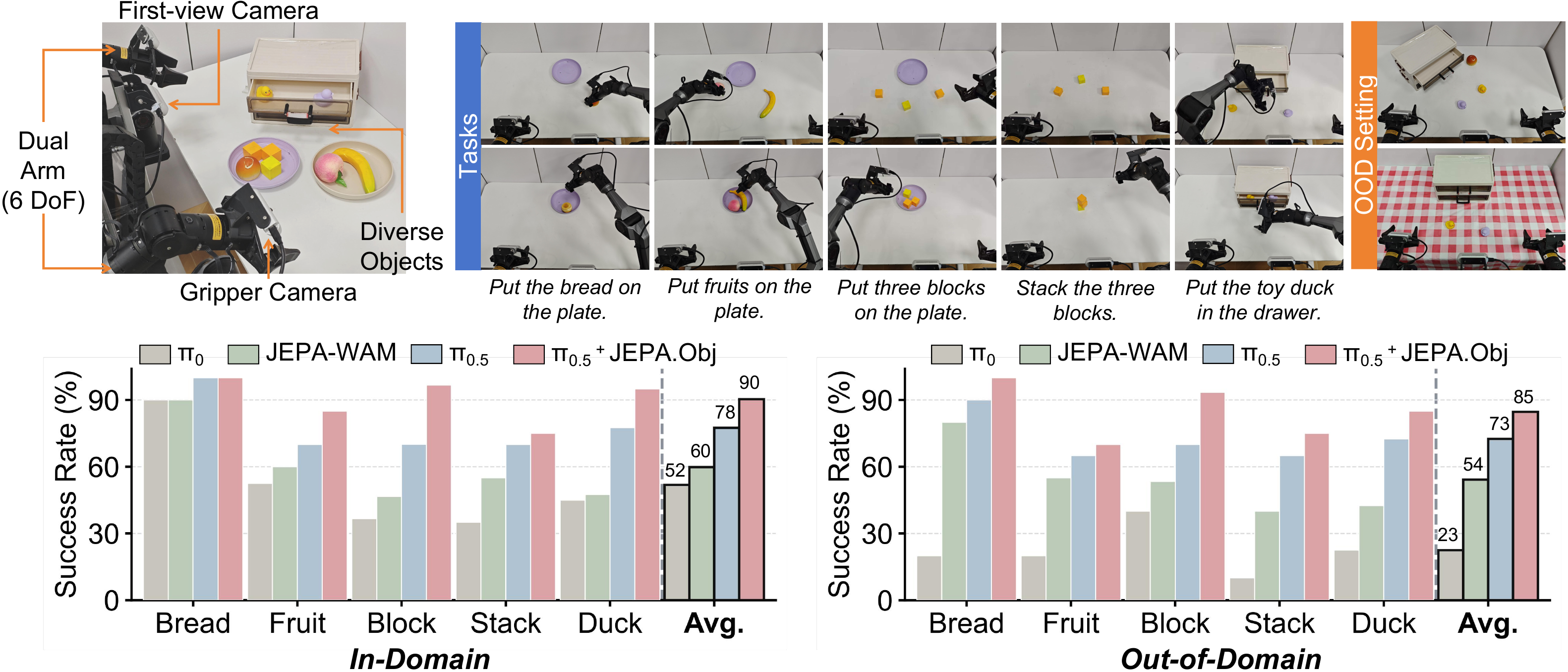}
    \caption{Real-world evaluation on five bimanual manipulation tasks under in-distribution (ID) and out-of-distribution (OOD) settings. We compare JEPA-WAM with VLA baselines and evaluate the effect of transition supervision on pretrained $\pi_{0.5}$.}
    \label{fig:real}
\end{figure*}

We conduct controlled ablations on LIBERO-Plus to analyze three key design choices of JEPA-WAM: the visual representation, the transition target, and its interaction with action generation.
All variants follow the same policy-training setup, with category-wise results reported in Table~\ref{tab:libero-plus-ablation}. Additional implementation details are provided in Appendix~C.

\paragraph{What visual representation should be used?}

JEPA-WAM is built in the pretrained V-JEPA representation space, which may contribute to robustness independently of transition prediction. To isolate this effect, we disable transition prediction and compare V-JEPA with DINOv2+SigLIP under the same setup.
As shown in Table~\ref{tab:libero-plus-ablation} (a, b), V-JEPA improves the average success rate from 73.2\% to 77.0\%, with larger gains under camera, lighting, and noise perturbations.
This result shows that the V-JEPA latent space itself provides a strong foundation for OOD policy learning.

\paragraph{What transition target should be predicted?}
We first ask whether the prediction target should represent the absolute future or the temporal relation between current and future observations.
We replace the joint current--future target with the future representation alone.
As shown in Table~\ref{tab:libero-plus-ablation} (c), the future-only variant achieves 77.3\%, compared with 79.2\% for the joint target.
This supports jointly representing both temporal endpoints: the future-only target specifies the resulting state, whereas the joint target makes their temporal relation directly available in the prediction target.

To better understand what is captured by the joint representation, we further compare it with future-only and endpoint-difference representations using frozen probes.
The probes show that, beyond the resulting state or the endpoint difference, the joint target preserves additional information about the temporal relation between the two observations, and that this temporal structure generalizes beyond the probe conditions seen during training.
Full protocols and results are provided in Appendix~C.

\paragraph{Should spatial structure be preserved?}
We next examine whether the spatial structure of the transition target should be preserved during supervision.
We apply an iREPA-style convolutional transformation to the same joint current--future target~\citep{singh2025matters}.
As shown in Table~\ref{tab:libero-plus-ablation} (d), the variant achieves 74.7\%, compared with 79.2\% for JEPA-WAM.
This supports preserving the original patch-level representation for transition supervision.
The convolutional transformation locally mixes neighboring features, which may weaken spatial correspondence and discard fine-grained details.

\paragraph{How should transition modeling interact with action generation?}

We first examine where transition supervision should be applied.
As shown in Table~\ref{tab:libero-plus-ablation} (e), using an intermediate layer in the Lower-16 variant reduces performance from 79.2\% to 76.5\%, suggesting that directly supervising the final shared predictor is more effective than using an intermediate auxiliary signal.

We further examine how the shared predictor should condition the action expert.
The Full-hidden variant removes the action placeholders and conditions the action expert on all last hidden states, reducing performance from 79.2\% to 73.1\% (Table~\ref{tab:libero-plus-ablation} (f)).
This suggests that directly sharing transition representations may cause interference between transition and action objectives, whereas action placeholders provide a dedicated action readout while preserving transition supervision on the shared predictor.

\subsection{Real-World Experiments}
\label{sec:real-world}

\paragraph{Setup.}
Figure~\ref{fig:real} shows the real-world platform, task examples, and ID/OOD evaluation settings.
We evaluate five bimanual manipulation tasks covering multi-object placement, long-horizon manipulation, and fine-grained spatial control.
OOD settings introduce changes in background or initial object configurations without additional fine-tuning.
Detailed platform, task, and evaluation protocols are provided in the Appendix.

\paragraph{Results.}
As shown in Figure~\ref{fig:real}, JEPA-WAM achieves an average score of 59.8\% under ID conditions and 54.2\% under OOD shifts, compared with 51.8\% and 22.5\% for the $\pi_0$.
In particular, JEPA-WAM maintains strong performance under changes in scene appearance and object configurations, consistent with the scene-level generalization observed in simulation.
For the pretrained $\pi_{0.5}$, instantiating the same transition supervision improves the average score from 77.5\% to 90.3\% under ID conditions and from 72.5\% to 84.7\% under OOD shifts.
These results show that JEPA-WAM generalizes well in real-world experiment, while the same transition supervision further improves both ID and OOD performance of a pretrained VLA policy. Complete per-rollout records are provided in Appendix~E.

\section{Conclusion}
We introduced JEPA-WAM, a latent world action model built in a pretrained V-JEPA representation space that couples latent transition modeling with continuous action generation through a shared predictor. JEPA-WAM uses a spatially structured joint current--future target to capture temporal relations and preserve patch-level structure, without requiring reconstruction of a unique future observation. The transition supervision emphasizes localized changes, object relations, and spatial reconfiguration, and directly shapes the backbone from which task-conditioned action representations are extracted. The same formulation can also be instantiated in pretrained VLA policies. Experiments across simulation and real-world manipulation demonstrate strong generalization under visual and spatial distribution shifts.

\section{Limitations}
JEPA-WAM is designed to learn general visual temporal structure rather than reconstruct a task-specific future. By supervising the relation between current and future observations, it captures shared patterns such as localized changes and evolving object and spatial relations that are largely independent of language. While such supervision can be broadly reused across tasks, it may be less expressive when the same observation leads to substantially different transitions under different instructions. Incorporating language-conditioned or multimodal transition targets is therefore a promising direction for future work.

\clearpage
\bibliography{aaai2027}

@inproceedings{assran2023ijepa,
  title={Self-Supervised Learning from Images with a Joint-Embedding Predictive Architecture},
  author={Assran, Mahmoud and Duval, Quentin and Misra, Ishan and Bojanowski, Piotr and Vincent, Pascal and Rabbat, Michael and LeCun, Yann and Ballas, Nicolas},
  booktitle={Proceedings of the IEEE/CVF Conference on Computer Vision and Pattern Recognition},
  pages={15619--15629},
  year={2023},
  url={https://openaccess.thecvf.com/content/CVPR2023/html/Assran_Self-Supervised_Learning_From_Images_With_a_Joint-Embedding_Predictive_Architecture_CVPR_2023_paper.html}
}

@article{bardes2024vjepa,
  title={Revisiting Feature Prediction for Learning Visual Representations from Video},
  author={Bardes, Adrien and Garrido, Quentin and Ponce, Jean and Chen, Xinlei and Rabbat, Michael and LeCun, Yann and Assran, Mahmoud and Ballas, Nicolas},
  year={2024},
  journal      = {Trans. Mach. Learn. Res.},
  volume       = {2024},
  url          = {https://openreview.net/forum?id=QaCCuDfBk2},
  bibsource    = {dblp computer science bibliography, https://dblp.org}
}

@misc{assran2025vjepa2,
  title={{V-JEPA 2}: Self-Supervised Video Models Enable Understanding, Prediction and Planning},
  author={Assran, Mido and Bardes, Adrien and Fan, David and Garrido, Quentin and Komeili, Mojtaba and Howes, Russell and Muckley, Matthew and Rizvi, Ammar and Roberts, Claire and Sinha, Koustuv and Zholus, Artem and others},
  year={2025},
  eprint={2506.09985},
  archivePrefix={arXiv},
  primaryClass={cs.AI},
  url={https://arxiv.org/abs/2506.09985}
}

@inproceedings{kim2024openvla,
  title={{OpenVLA}: An Open-Source Vision-Language-Action Model},
  author={Kim, Moo Jin and Pertsch, Karl and Karamcheti, Siddharth and Xiao, Ted and Balakrishna, Ashwin and Nair, Suraj and Rafailov, Rafael and Foster, Ethan and Lam, Grace and Sanketi, Pannag and others},
  booktitle    = {Conference on Robot Learning, 6-9 November 2024, Munich, Germany},
  series       = {Proceedings of Machine Learning Research},
  volume       = {270},
  pages        = {2679--2713},
  publisher    = {{PMLR}},
  year         = {2024},
  url          = {https://proceedings.mlr.press/v270/kim25c.html},
  bibsource    = {dblp computer science bibliography, https://dblp.org}
}

@misc{black2024pi_0,
  title={{${\pi}_0$}: A Vision-Language-Action Flow Model for General Robot Control},
  author={Black, Kevin and Brown, Noah and Driess, Danny and Esmail, Adnan and Equi, Michael and Finn, Chelsea and Fusai, Niccolo and Groom, Lachy and Hausman, Karol and Ichter, Brian and others},
  year={2024},
  eprint={2410.24164},
  archivePrefix={arXiv},
  primaryClass={cs.LG},
  url={https://arxiv.org/abs/2410.24164}
}

@inproceedings{intelligence2025pi_,
  title={{${\pi}_{0.5}$}: A Vision-Language-Action Model with Open-World Generalization},
  author={{Physical Intelligence} and Black, Kevin and Brown, Noah and Darpinian, James and Dhabalia, Karan and Driess, Danny and Esmail, Adnan and Equi, Michael and Finn, Chelsea and Fusai, Niccolo and others},
  booktitle={9th Annual Conference on Robot Learning},
  year={2025}
}

@inproceedings{liu2025rdt,
  title={{RDT-1B}: A Diffusion Foundation Model for Bimanual Manipulation},
  author={Liu, Songming and Wu, Lingxuan and Li, Bangguo and Tan, Hengkai and Chen, Huayu and Wang, Zhengyi and Xu, Ke and Su, Hang and Zhu, Jun},
  booktitle={International Conference on Learning Representations},
  year={2025},
  url={https://arxiv.org/abs/2410.07864}
}

@misc{li2026training,
  title={Training Vision-Language-Action Models with Dense Embodied Chain-of-Thought Supervision},
  author={Li, Haoyang and Li, Guanlin and Feng, Youhe and Zhao, Chen and Wang, Zhuoran and Li, Yang and Wei, Qizhe and Bao, Shifeng and Shen, Haitao and Zhao, Yihan and others},
  year={2026},
  eprint={2606.30552},
  archivePrefix={arXiv},
  primaryClass={cs.RO},
  url={https://arxiv.org/abs/2606.30552}
}

@misc{ye2026world,
  title={World Action Models Are Zero-Shot Policies},
  author={Ye, Seonghyeon and Ge, Yunhao and Zheng, Kaiyuan and Gao, Shenyuan and Yu, Sihyun and Kurian, George and Indupuru, Suneel and Tan, You Liang and Zhu, Chuning and Xiang, Jiannan and others},
  year={2026},
  eprint={2602.15922},
  archivePrefix={arXiv},
  primaryClass={cs.RO},
  url={https://arxiv.org/abs/2602.15922}
}

@misc{li2026causal,
  title={Causal World Modeling for Robot Control},
  author={Li, Lin and Zhang, Qihang and Luo, Yiming and Yang, Shuai and Wang, Ruilin and Han, Fei and Yu, Mingrui and Gao, Zelin and Xue, Nan and Zhu, Xing and others},
  year={2026},
  eprint={2601.21998},
  archivePrefix={arXiv},
  primaryClass={cs.RO},
  url={https://arxiv.org/abs/2601.21998}
}

@misc{cen2025worldvla,
  title={{WorldVLA}: Towards Autoregressive Action World Model},
  author={Cen, Jun and Yu, Chaohui and Yuan, Hangjie and Jiang, Yuming and Huang, Siteng and Guo, Jiayan and Li, Xin and Song, Yibing and Luo, Hao and Wang, Fan and others},
  year={2025},
  eprint={2506.21539},
  archivePrefix={arXiv},
  primaryClass={cs.RO},
  url={https://arxiv.org/abs/2506.21539}
}

@misc{chen2026lawam,
  title={{LaWAM}: Latent World Action Models for Efficient Dynamics-Aware Robot Policies},
  author={Chen, Jialei and Wang, Kai and Chen, Kang and Chen, Shuaihang and Gao, Feng and Tang, Wenhao and Li, Zhiyuan and Liu, Weilin and Yao, Zhuyu and Li, Boxun and Xu, Yuanbo and Yu, Chao},
  year={2026},
  eprint={2606.15768},
  archivePrefix={arXiv},
  primaryClass={cs.RO},
  url={https://arxiv.org/abs/2606.15768}
}

@misc{li2026lightwam,
  title={{Light-WAM}: Efficient World Action Models with State-Fusion Action Decoding},
  author={Li, Ziang and Cheng, Dongzhou and Wang, Yibin and Wang, Shiyue and Xu, Xiaoyang and Weng, Lingxuan and Wang, Juan and Wang, Jiaqi},
  year={2026},
  eprint={2606.08242},
  archivePrefix={arXiv},
  primaryClass={cs.CV},
  url={https://arxiv.org/abs/2606.08242}
}

@misc{miao2026jepa,
  title={{JEPA-VLA}: Video Predictive Embedding Is Needed for VLA Models},
  author={Miao, Shangchen and Feng, Ningya and Wu, Jialong and Lin, Ye and He, Xu and Li, Dong and Long, Mingsheng},
  year={2026},
  eprint={2602.11832},
  archivePrefix={arXiv},
  primaryClass={cs.RO},
  url={https://arxiv.org/abs/2602.11832}
}

@misc{starvla,
  title={{StarVLA}: A Lego-Like Codebase for Vision-Language-Action Model Developing},
  author={{StarVLA Community}},
  year={2026},
  eprint={2604.05014},
  archivePrefix={arXiv},
  primaryClass={cs.RO},
  url={https://arxiv.org/abs/2604.05014}
}

@inproceedings{prismaticvlm,
  title={Prismatic {VLMs}: Investigating the Design Space of Visually-Conditioned Language Models},
  author={Karamcheti, Siddharth and Nair, Suraj and Balakrishna, Ashwin and Liang, Percy and Kollar, Thomas and Sadigh, Dorsa},
  booktitle={Forty-first International Conference on Machine Learning, {ICML} 2024,
             Vienna, Austria, July 21--27, 2024},
  series={Proceedings of Machine Learning Research},
  volume={235},
  pages={23123--23144},
  year={2024}
}

@article{chi2024diffusionpolicyvisuomotorpolicy,
  title={Diffusion Policy: Visuomotor Policy Learning via Action Diffusion},
  author={Chi, Cheng and Xu, Zhenjia and Feng, Siyuan and Cousineau, Eric and Du, Yilun and Burchfiel, Benjamin and Tedrake, Russ and Song, Shuran},
  journal={The International Journal of Robotics Research},
  volume={44},
  number={10-11},
  pages={1684--1704},
  year={2025},
  publisher={Sage Publications Sage UK: London, England}
}

@misc{fei2025liberoplus,
  title={{LIBERO-Plus}: In-Depth Robustness Analysis of Vision-Language-Action Models},
  author={Fei, Senyu and Wang, Siyin and Shi, Junhao and Dai, Zihao and Cai, Jikun and Qian, Pengfang and Ji, Li and He, Xinzhe and Zhang, Shiduo and Fei, Zhaoye and others},
  year={2025},
  eprint={2510.13626},
  archivePrefix={arXiv},
  primaryClass={cs.RO},
  url={https://arxiv.org/abs/2510.13626}
}

@inproceedings{liu2023libero,
  title={{LIBERO}: Benchmarking Knowledge Transfer for Lifelong Robot Learning},
  author={Liu, Bo and Zhu, Yifeng and Gao, Chongkai and Feng, Yihao and Liu, Qiang and Zhu, Yuke and Stone, Peter},
  booktitle={Advances in Neural Information Processing Systems},
  volume={36},
  pages={44776--44791},
  year={2023}
}

@misc{zhong2026noiseintentanchoringgenerative,
  title={From Noise to Intent: Anchoring Generative {VLA} Policies with Residual Bridges},
  author={Zhong, Yiming and He, Yaoyu and Yang, Zemin and Tian, Pengfei and Huang, Yifan and Huang, Qingqiu and Zhu, Xinge and Ma, Yuexin},
  year={2026},
  eprint={2604.21391},
  archivePrefix={arXiv},
  primaryClass={cs.RO},
  url={https://arxiv.org/abs/2604.21391}
}

@inproceedings{wang2025vlaadaptereffectiveparadigmtinyscale,
  title={{VLA-Adapter}: An Effective Paradigm for Tiny-Scale Vision-Language-Action Models},
  author={Wang, Yihao and Ding, Pengxiang and Li, Lingxiao and Cui, Can and Ge, Zirui and Tong, Xinyang and Song, Wenxuan and Zhao, Han and Zhao, Wei and Hou, Pengxu and others},
  booktitle={Proceedings of the AAAI conference on artificial intelligence},
  volume={40},
  pages={18638--18646},
  year={2026}
}

@misc{sun2026vlajepa,
  title={{VLA-JEPA}: Enhancing Vision-Language-Action Model with Latent World Model},
  author={Sun, Jingwen and Zhang, Wenyao and Qi, Zekun and Ren, Shaojie and Liu, Zezhi and Zhu, Hanxin and Sun, Guangzhong and Jin, Xin and Chen, Zhibo},
  year={2026},
  eprint={2602.10098},
  archivePrefix={arXiv},
  primaryClass={cs.RO},
  url={https://arxiv.org/abs/2602.10098}
}

@misc{chen2025robotwin,
  title={{RoboTwin 2.0}: A Scalable Data Generator and Benchmark with Strong Domain Randomization for Robust Bimanual Robotic Manipulation},
  author={Chen, Tianxing and Chen, Zanxin and Chen, Baijun and Cai, Zijian and Liu, Yibin and Li, Zixuan and Liang, Qiwei and Lin, Xianliang and Ge, Yiheng and Gu, Zhenyu and others},
  year={2025},
  eprint={2506.18088},
  archivePrefix={arXiv},
  primaryClass={cs.RO},
  url={https://arxiv.org/abs/2506.18088}
}

@misc{luo2026rovlamulticonsistencyconstraintsrobust,
  title={{RoVLA}: Multi-Consistency Constraints for Robust Vision-Language-Action Models},
  author={Luo, Jingzhou and Wen, Yifan and Bai, Yongjie and Song, Xinshuai and Liu, Yang and Lin, Liang},
  year={2026},
  eprint={2605.19678},
  archivePrefix={arXiv},
  primaryClass={cs.RO},
  url={https://arxiv.org/abs/2605.19678}
}

@misc{zheng2026pokevla,
  title={{PokeVLA}: Empowering Pocket-Sized Vision-Language-Action Model with Comprehensive World Knowledge Guidance},
  author={Zheng, Yupeng and Li, Xiang and Gu, Songen and Zheng, Yuhang and Tian, Shuai and Li, Weize and Wang, Linbo and Fei, Senyu and Li, Pengfei and Gao, Yinfeng and others},
  year={2026},
  eprint={2604.20834},
  archivePrefix={arXiv},
  primaryClass={cs.RO},
  url={https://arxiv.org/abs/2604.20834}
}

@misc{ye2026gigaworld,
  title={GigaWorld-Policy: An Efficient Action-Centered World--Action Model},
  author={Ye, Angen and Wang, Boyuan and Ni, Chaojun and Huang, Guan and Zhao, Guosheng and Li, Hao and Li, Hengtao and Li, Jie and Lv, Jindi and Liu, Jingyu and others},
  journal={arXiv preprint arXiv:2603.17240},
  year={2026}
}

@misc{mur2026v,
  title={V-jepa 2.1: Unlocking dense features in video self-supervised learning},
  author={Mur-Labadia, Lorenzo and Muckley, Matthew and Bar, Amir and Assran, Mido and Sinha, Koustuv and Rabbat, Mike and LeCun, Yann and Ballas, Nicolas and Bardes, Adrien},
  journal={arXiv preprint arXiv:2603.14482},
  year={2026}
}

@misc{yang2026abotm0,
  title   = {{ABot-M0}: {VLA} Foundation Model for Robotic Manipulation with Action Manifold Learning},
  author  = {Yang, Yandan and Zeng, Shuang and Lin, Tong and Chang, Xinyuan
             and Qi, Dekang and Xiao, Junjin and Liu, Haoyun and Chen, Ronghan
             and Chen, Yuzhi and Huo, Dongjie and Xiong, Feng and Wei, Xing
             and Ma, Zhiheng and Xu, Mu},
  journal = {arXiv preprint arXiv:2602.11236},
  year    = {2026}
}

@misc{kim2026cosmospolicy,
  title   = {Cosmos Policy: Fine-Tuning Video Models for Visuomotor Control and Planning},
  author  = {Kim, Moo Jin and Gao, Yihuai and Lin, Tsung-Yi and Lin, Yen-Chen
             and Ge, Yunhao and Lam, Grace and Liang, Percy and Song, Shuran
             and Liu, Ming-Yu and Finn, Chelsea and Gu, Jinwei},
  journal = {arXiv preprint arXiv:2601.16163},
  year    = {2026}
}

@misc{luo2026beingh07,
  title   = {{Being-H0.7}: A Latent World-Action Model from Egocentric Videos},
  author  = {Luo, Hao and Zhang, Wanpeng and Feng, Yicheng and Zheng, Sipeng
             and Xu, Haiweng and Xu, Chaoyi and Xi, Ziheng and Fu, Yuhui
             and Lu, Zongqing},
  journal = {arXiv preprint arXiv:2605.00078},
  year    = {2026}
}

@inproceedings{zheng2025flare,
  title={Flare: Robot learning with implicit world modeling},
  author={Zheng, Ruijie and Wang, Jing and Reed, Scott and Bjorck, Johan and Fang, Yu and Hu, Fengyuan and Jang, Joel and Kundalia, Kaushil and Lin, Zongyu and Magne, Loic and others},
  booktitle = 	 {Proceedings of The 9th Conference on Robot Learning},
  pages = 	 {3952--3971},
  year = 	 {2025},
  volume = 	 {305},
  series = 	 {Proceedings of Machine Learning Research},
  publisher =    {PMLR},
}

@misc{zhao2026frappe,
  title={Frappe: Infusing world modeling into generalist policies via multiple future representation alignment},
  author={Zhao, Han and Wang, Jingbo and Song, Wenxuan and Chen, Shuai and Liu, Yang and Wang, Yan and Li, Haoang and Wang, Donglin},
  journal={arXiv preprint arXiv:2602.17259},
  year={2026}
}

@misc{yuan2026fastwam,
  title={Fast-WAM: Do World Action Models Need Test-time Future Imagination?},
  author={Tianyuan Yuan and Zibin Dong and Yicheng Liu and Hang Zhao},
  journal={arXiv preprint arXiv:2603.16666},
  year={2026},
  url={https://arxiv.org/abs/2603.16666}
}

@misc{ma2026dit4dit,
  title={Dit4dit: Jointly modeling video dynamics and actions for generalizable robot control},
  author={Ma, Teli and Zheng, Jia and Wang, Zifan and Jiang, Chunli and Cui, Andy and Liang, Junwei and Yang, Shuo},
  journal={arXiv preprint arXiv:2603.10448},
  year={2026}
}

@misc{bi2025motus,
  title={Motus: A unified latent action world model},
  author={Bi, Hongzhe and Tan, Hengkai and Xie, Shenghao and Wang, Zeyuan and Huang, Shuhe and Liu, Haitian and Zhao, Ruowen and Feng, Yao and Xiang, Chendong and Rong, Yinze and others},
  journal={arXiv preprint arXiv:2512.13030},
  year={2025}
}

@misc{singh2025matters,
  title={What matters for Representation Alignment: Global Information or Spatial Structure?},
  author={Singh, Jaskirat and Leng, Xingjian and Wu, Zongze and Zheng, Liang and Zhang, Richard and Shechtman, Eli and Xie, Saining},
  journal={arXiv preprint arXiv:2512.10794},
  year={2025}
}

@article{li2025jit,
  title   = {Back to Basics: Let Denoising Generative Models Denoise},
  author  = {Li, Tianhong and He, Kaiming},
  journal = {arXiv preprint arXiv:2511.13720},
  year    = {2025}
}

@article{jeon2026vision,
  title   = {Vision-aligned Latent Reasoning for Multi-modal Large Language Model},
  author  = {Jeon, Byungwoo and Jeong, Yoonwoo and Lee, Hyunseok and Cho, Minsu and Shin, Jinwoo},
  journal = {arXiv preprint arXiv:2602.04476},
  year    = {2026}
}

\clearpage

\onecolumn
\appendix

\section{Implementation Details}
\label{app:implementation}

\subsection{JEPA-WAM Architecture}

JEPA-WAM uses a frozen V-JEPA 2.1 ViT-L/16 encoder with an input resolution of $384\times384$.
For each camera view, the encoder produces a $24\times24$ grid of 1024-dimensional patch features.
Multi-view observations are encoded independently and concatenated in a fixed camera order without spatial pooling.

A two-layer visual projector $P_{\mathrm{vis}}$ maps the V-JEPA features into the predictor hidden space,
\[
1024 \rightarrow 896 \rightarrow 896,
\]
with GELU activation.
The shared predictor $F_\theta$ is initialized from Qwen2.5-0.5B with hidden dimension 896.
During robot-policy training, the base predictor is adapted using LoRA with rank 32, scaling factor 64, and dropout 0.1.

For latent transition prediction, the final predictor hidden states at the visual-token positions are mapped back to the V-JEPA representation space by a token-wise MLP,
\[
896 \rightarrow 2048 \rightarrow 1024,
\]
with GELU activation.
The resulting features preserve the camera and patch-token ordering of the current visual representation and are supervised by patch-wise cosine distance to the frozen joint current--future target.

For action generation, we append 64 dedicated action placeholders to the predictor sequence and use their final hidden states as the action-conditioning representation $C_t$.
The action expert is a 16-layer DiT-L flow-matching model.
It additionally receives the proprioceptive state and 32 learnable future tokens, and predicts a continuous action chunk conditioned on $C_t$.

\paragraph{Joint current--future target.}
For each view $v$, the current observation $O_t^v$ and the future observation $O_{t+\delta}^v$ are stacked along the temporal dimension and jointly encoded by the frozen V-JEPA encoder.
Because V-JEPA uses a temporal tubelet size of two, the two-frame input produces one temporal token per spatial location and therefore retains the same $24\times24$ spatial grid as the current representation.
The resulting target is detached from the computation graph and preserves the same view and patch ordering as the current visual tokens.

For LIBERO, we use one primary and one wrist-camera view, an action horizon of $H=8$, and a temporal target offset of $\delta=31$.
Near the end of a trajectory, the future observation is clipped to the last available frame.

For RoboTwin 2.0, we use one primary and two wrist-camera views, an action horizon of $H=50$, and a temporal target offset of $\delta=50$.
Near the end of a trajectory, the future observation is clipped to the last available frame.

\subsection{Vision--Language Initialization}

Before robot-policy training, we initialize the vision-language interface following the single-stage finetuning setup of Prismatic~\citep{prismaticvlm}. The V-JEPA 2.1 encoder is kept frozen, while the visual projector and the full Qwen2.5-0.5B language-model backbone are jointly finetuned, without a separate projector-alignment stage.
We train on the LLaVA v1.5 using autoregressive language modeling on assistant responses.

We train for two epochs using AdamW with a learning rate of $2\times10^{-5}$, weight decay 0.1, cosine decay, 3\% warmup, BF16 precision, and a global batch size of 128.
During subsequent robot-policy training, the V-JEPA encoder, visual projector, and base Qwen weights are frozen.
Only the Qwen LoRA adapters, transition prediction head, and action expert are optimized.

\subsection{Robot Policy Training}

During robot-policy training, JEPA-WAM jointly optimizes action generation and latent transition prediction,
\begin{equation}
\mathcal{L}
=
\mathcal{L}_{\mathrm{act}}
+
\lambda_{\mathrm{wm}}\mathcal{L}_{\mathrm{wm}},
\qquad
\lambda_{\mathrm{wm}}=0.5.
\end{equation}
The transition loss is the mean patch-wise cosine distance between the predicted representation and the detached joint current--future V-JEPA target.

For action generation, we use conditional flow matching as described in the main paper.
Given an action chunk $a$, Gaussian noise $\epsilon\sim\mathcal{N}(0,I)$, and a flow time $\tau$ sampled using a Beta-based schedule with parameters $\alpha=1.5$ and $\beta=1.0$, we construct
\begin{equation}
a_\tau=(1-\tau)\epsilon+\tau a,
\end{equation}
and train the action expert with velocity prediction,
\begin{equation}
\mathcal{L}_{\mathrm{act}}
=
\mathbb{E}_{\epsilon,\tau}
\left[
\left\|
A_\psi(a_\tau,\tau,s_t,C_t)
-
(a-\epsilon)
\right\|_2^2
\right].
\end{equation}

For the main LIBERO training, we jointly train on the four standard suites using the primary and wrist-camera observations.
We use AdamW with a peak learning rate of $2\times10^{-4}$, cosine decay to $10^{-5}$, 3\% warmup, zero weight decay, and gradient clipping at 1.0.
Training uses BF16 and FSDP on eight GPUs with a global batch size of 128, and runs for 60K optimization steps.
Benchmark-specific observation and action configurations are provided in Appendix~B.

\subsection{Transfer to Pretrained VLA Policies}

Inspired by VaLR~\citep{jeon2026vision}, which introduces learnable latent tokens and aligns their hidden representations with dense visual features, we instantiate joint current--future supervision in the pretrained $\pi_{0.5}$ policy through a lightweight future tokens.
Specifically, we append 64 learnable future tokens to the VLM prefix while preserving the policy's original perception and action-generation pathways.
We use their output hidden states
\[
R_t \in \mathbb{R}^{64\times2048}
\]
for transition prediction.
The 64 tokens are arranged as an $8\times8$ coarse spatial grid,
\[
\widetilde{R}_t
=
\operatorname{Reshape}(R_t)
\in
\mathbb{R}^{8\times8\times2048}.
\]

A lightweight spatial prediction head applies LayerNorm followed by an MLP,
\[
2048 \rightarrow 2048 \rightarrow 1408,
\]
with GELU activation.
The resulting $8\times8$ feature map is bilinearly upsampled to $24\times24$ and aligned patch-wise with a frozen V-JEPA 2.1 ViT-G joint current--future target.
The target is constructed from the current frame and the frame at offset $\delta$, with the future frame clipped to the final frame near trajectory boundaries.

We optimize the original flow-matching action loss together with the auxiliary transition loss.
The transition-loss weight is linearly warmed up during the first 1K optimization steps to $\lambda_{\mathrm{wm}}=0.1$.
The future tokens can attend to the original image and language prefix, while the action tokens are prevented from attending to these newly introduced queries.
Thus, the auxiliary branch shapes the shared VLA backbone through transition supervision without introducing additional future-token conditioning into the original action pathway.

\subsection{Deployment}

JEPA-WAM requires only the current observation and language instruction at deployment.
Future observations are not required, and neither joint target encoding nor the transition prediction head is executed during inference.
The deployed policy therefore consists of the current-frame V-JEPA encoder, visual projector, shared predictor, and action expert.
Starting from Gaussian noise, the action expert generates each action chunk using four Euler integration steps of the learned flow.

For $\pi_{0.5}$ with transition supervision, the V-JEPA teacher, transition targets, spatial prediction head, and auxiliary loss are used only during training.
The learned future tokens remain part of the VLM prefix at inference, but the original action tokens remain masked from attending to them.
Thus, no predicted future representation is explicitly provided to the action expert, and action generation follows the original $\pi_{0.5}$ pathway.

\clearpage
\section{Experimental Details}
\label{app:experimental-details}

\subsection{LIBERO and LIBERO-Plus}

\paragraph{Dataset and evaluation.}
We evaluate on LIBERO~\citep{liu2023libero} and LIBERO-Plus~\citep{fei2025liberoplus}.
For LIBERO, we jointly train on the four standard task suites and evaluate the resulting policy on the corresponding test tasks.
For LIBERO-Plus, we directly transfer the policy trained on LIBERO demonstrations to the perturbed environments without additional fine-tuning.
We report success rates following the standard evaluation protocol.

\paragraph{Training configuration.}
The policy receives images from the primary camera and wrist camera together with language instructions and proprioceptive states.
The action space contains 7-dimensional robot actions with an action chunk length of 8.
For JEPA-WAM, we construct the transition target using the observation pair separated by $\delta=31$ frames.

The LIBERO model is trained with AdamW using a peak learning rate of $2\times10^{-4}$ and cosine learning-rate decay.
We use a 3\% warmup ratio, zero weight decay, gradient clipping with a maximum norm of 1.0, and BF16 precision.
Training is performed on 8 GPUs with a global batch size of 128 for 60K optimization steps.

\subsection{RoboTwin 2.0}
\label{appendix_b_2}

\paragraph{Dataset and evaluation.}
We evaluate on RoboTwin 2.0~\citep{chen2025robotwin} using 20 manipulation tasks, organized into two groups of 10 tasks.
For each task, the policy is trained only on demonstrations from the Clean setting.
The same trained policy is then evaluated on both the Clean and Random settings, where Random introduces changes to the scene and object configurations.
We report the average task success rate over all 20 tasks.

\paragraph{Observation and action configuration.}
RoboTwin provides one external camera and two wrist cameras. The three views are independently encoded by the frozen V-JEPA encoder, and the resulting visual tokens are concatenated in a fixed camera order without spatial pooling. The policy predicts 14-dimensional bimanual actions with an action horizon of 50 steps.

For RoboTwin, we use $x$-prediction in the flow-matching action expert, directly predicting the clean action trajectory from the noisy trajectory.
We found this parameterization to provide more stable training for the longer bimanual action chunks.
This choice is also motivated by JiT~\citep{li2025jit}, which advocates directly predicting clean data rather than noised quantities.

Complete task-wise results for all 20 tasks are provided in Appendix~D.

\subsection{Real-World Setup}

\paragraph{Platform and data collection.}
We evaluate on a bimanual AgileX Cobot Magic platform with two 6-DoF arms and grippers.
Visual observations are provided by one global camera and two wrist cameras.
We consider five manipulation tasks: placing bread on a plate, placing peaches and bananas on a plate, placing three blocks on a plate, stacking three blocks, and placing toy ducks in a drawer.
For each task, we collect 100 demonstrations for policy training.

\paragraph{Evaluation protocol.}
Each policy is evaluated under both in-distribution (ID) and out-of-distribution (OOD) settings.
The ID setting follows the environment used during data collection.
For OOD evaluation, we introduce changes in the background or initial object configurations. The task objective and language instruction remain unchanged, and no additional fine-tuning is performed.

We conduct 10 rollouts for each task and evaluation setting.
Because several tasks contain multiple subgoals, we report normalized task-completion scores in $[0,1]$ rather than only binary success.
The task-specific scoring criteria are summarized in Table~\ref{tab:real-world-tasks}.

\begin{table}[!htbp]
\centering
\small
\begin{tabular}{@{}p{0.31\textwidth}p{0.43\textwidth}p{0.18\textwidth}@{}}
\toprule
Task & Evaluation objective & Recorded scores \\
\midrule
Bread placement
& Place the target bread on the plate.
& $0,1$ \\
Fruit placement
& Place the target peaches and bananas on the plate with partial credit for completed subgoals.
& $0,.25,.50,.75,1$ \\
Three-block placement
& Place all three target blocks on the plate, with partial credit for completed placements.
& $0,.33,.67,1$ \\
Three-block stacking
& Form the three-block stack with partial credit for completed stacking relations.
& $0,.50,1$ \\
Drawer manipulation
& Open the drawer, place two target toy ducks, and close the drawer.
& $0,.25,.50,.75,1$ \\
\bottomrule
\end{tabular}
\caption{Real-world tasks and normalized completion-score definitions.}
\label{tab:real-world-tasks}
\end{table}
\clearpage
\section{Additional Analysis}

\subsection{Design Ablations}

We provide additional details for the controlled ablations in the main paper.
The \textit{V-JEPA only} variant removes transition prediction while retaining V-JEPA as the policy visual encoder.
The \textit{Future only} variant replaces the joint current--future target with the representation of the future observation alone.
The \textit{Endpoint difference} variant independently encodes the current and future observations and uses their feature difference,
\begin{equation}
Y^{\mathrm{diff}}_{t,t+\delta}
=
E_J(O_{t+\delta}) - E_J(O_t),
\end{equation}
as the prediction target.
The \textit{iREPA align.} variant replaces direct token-wise alignment with a per-view $3\times3$ convolutional transformation and spatial target normalization.
The \textit{Lower-16 align.} variant applies the same transition objective to the hidden states from the 16th predictor layer.
The \textit{Full hidden} variant removes the dedicated action placeholders and conditions the action expert on the complete final predictor hidden sequence.

To further distinguish joint encoding from explicitly differencing the two endpoints, we evaluate the endpoint-difference target under the same LIBERO-Plus setting.
Table~\ref{tab:target_construction} compares the three target constructions.

\begin{table}[!htbp]
\centering
\small
\setlength{\tabcolsep}{3.2pt}
\begin{tabular}{lcccccccc}
\toprule
Target & Cam. & Rob. & Lang. & Lit. & Back. & Noi. & Lay. & Avg. \\
\midrule
Endpoint difference
& 54.5 & 49.2 & 73.5 & 89.6 & 86.2 & 66.6 & 76.9 & 70.9 \\
Future only
& 75.1 & 47.1 & 69.6 & 96.0 & 93.4 & 81.5 & 78.4 & 77.3 \\
Joint current--future
& \textbf{79.2} & \textbf{59.2} & 68.2 & 93.3 & \textbf{94.6} & \textbf{83.6} & 76.1 & \textbf{79.2} \\
\bottomrule
\end{tabular}
\caption{LIBERO-Plus success rates (\%) for different transition-target constructions.}
\label{tab:target_construction}
\end{table}

The joint current--future target achieves 79.2\%, compared with 77.3\% for future-only prediction and 70.9\% for explicit endpoint differencing.
Thus, simply providing the resulting future state or explicitly subtracting independently encoded endpoints does not recover the benefit of joint encoding.
Together with the remaining ablations in the main paper, direct patch-level supervision improves the average by 4.5 points over the iREPA-style alignment, final-layer transition supervision improves by 2.7 points over the lower-layer variant, and dedicated action-conditioning representations improve by 6.1 points over conditioning on the full hidden sequence.

The particularly large degradation from joint encoding to endpoint differencing motivates a closer analysis of what information is represented by the joint target.

\subsection{Transition Information in the Joint Current--Future Target}

The policy ablations show that the joint current--future target provides more effective supervision than both future-only prediction and explicit endpoint differencing.
We next analyze the frozen target representations to determine whether joint encoding preserves transition information beyond first-order change between the two endpoints.
We consider three complementary probes: controlled temporal-gap decoding, generalization to unseen temporal gaps, and recovery of trajectory structure after removing endpoint displacement.

\paragraph{Representations and probe setup.}
For a current--future pair $(O_t,O_f)$, we define the pooled joint representation as
\begin{equation}
z_{\mathrm{joint}}
=
\operatorname{Pool}
\left(
E_J
\left(
\operatorname{Stack}_{\mathrm{time}}(O_t,O_f)
\right)
\right),
\end{equation}
and construct an explicit endpoint-difference baseline by independently encoding the two observations,
\begin{equation}
z_{\mathrm{diff}}
=
\operatorname{Pool}(E_J(O_f))
-
\operatorname{Pool}(E_J(O_t)).
\end{equation}

All representations are frozen and evaluated using the same ridge linear probe.
We use RoboTwin Clean-20 with 1,000 episodes from 20 tasks and the external camera.
For each task, 30 episodes are used for probe training, 10 for validation, and 10 for testing.
All splits are episode-disjoint.
Probe regularization is selected only on the validation set, and uncertainty is estimated using 1,000 paired episode-level bootstrap resamples.

\paragraph{Controlled temporal-gap decoding.}
We first ask whether the joint target represents the temporal relation between two observations rather than merely their endpoint content.
For each fixed future anchor $O_f$, we construct
\begin{equation}
(O_f,O_f),~
(O_{f-10},O_f),~
(O_{f-20},O_f),~
(O_{f-30},O_f),~
(O_{f-40},O_f),~
(O_{f-50},O_f),
\end{equation}
corresponding to temporal gaps
$g\in\{0,10,20,30,40,50\}$.
The future observation is therefore identical across all six classes, while only the current observation and its relation to the future change.
We use two future anchors per episode, yielding 12,000 pairs in total.

A six-way linear classifier predicts the temporal gap from each frozen representation.
Future-only features provide a sanity check because they are identical across the six gap values for a fixed anchor.
Current-only features may exploit trajectory phase, while endpoint differencing provides the stronger comparison because it has access to both endpoints but represents their relation through explicit subtraction.

\begin{table}[!htbp]
\centering
\small
\begin{tabular}{lcc}
\toprule
Representation & Accuracy (\%) $\uparrow$ & 95\% CI \\
\midrule
Future only & 16.7 & 16.7--16.7 \\
Current only & 44.6 & 42.5--47.0 \\
Endpoint difference & 47.0 & 45.0--49.1 \\
Joint target & \textbf{67.2} & \textbf{65.3--69.1} \\
\bottomrule
\end{tabular}
\caption{Controlled temporal-gap decoding with the future observation fixed. Chance accuracy is 16.7\%.}
\label{tab:gap_decoding}
\end{table}

The joint target reaches 67.2\% accuracy, substantially above endpoint differencing at 47.0\%.
The paired improvement is 20.1 accuracy points, with a 95\% bootstrap confidence interval of $[18.3,21.9]$.
Since the future observation is controlled and endpoint differencing already has access to both observations, this result shows that joint encoding makes their temporal relation substantially more accessible than independent endpoint subtraction.

\begin{figure}[!t]
\centering
\includegraphics[width=\linewidth]{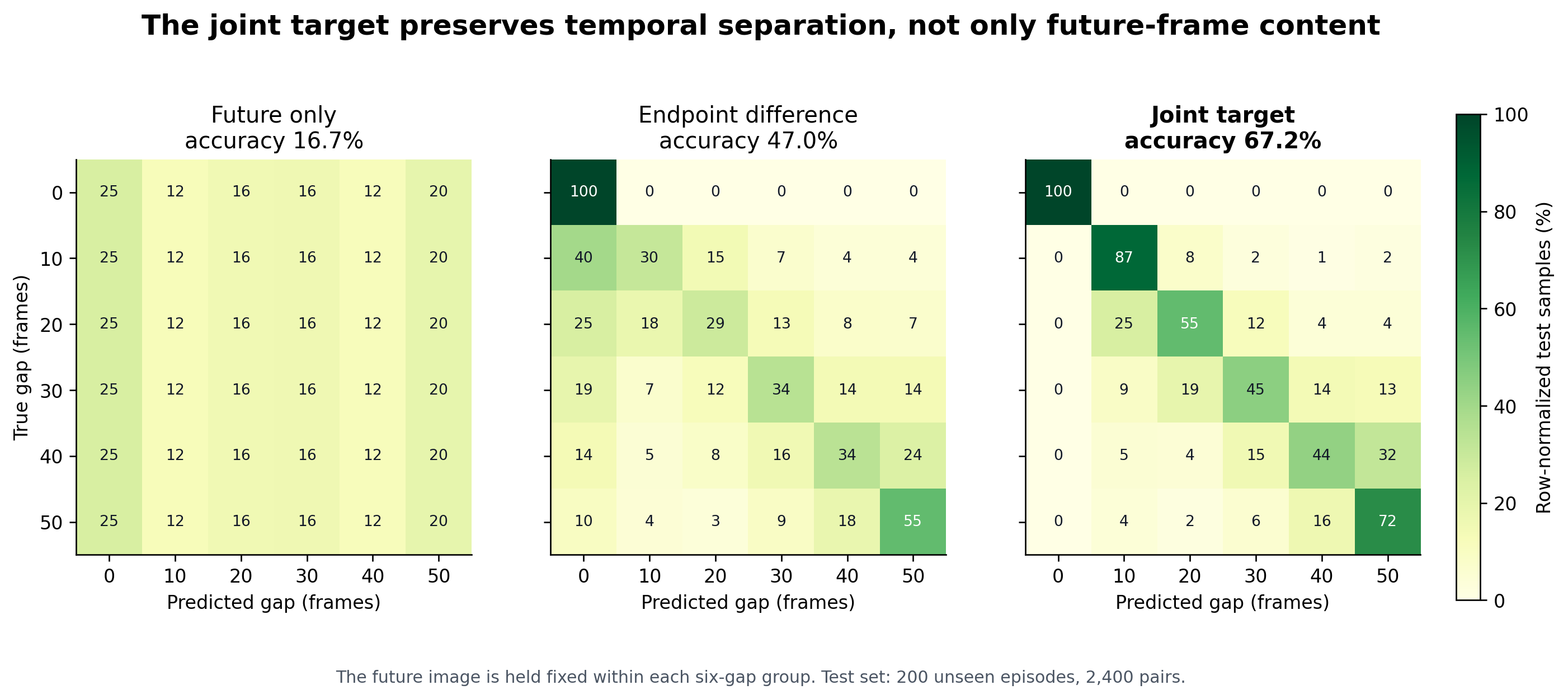}
\caption{
Temporal-gap decoding with a fixed future observation.
Rows denote true gaps and columns denote predicted gaps.
Future-only features are uninformative by construction, endpoint differencing captures partial temporal ordering, and the joint target exhibits a substantially clearer diagonal structure.
}
\label{fig:gap_confusion}
\end{figure}

\paragraph{Generalization to unseen temporal gaps.}
Gap classification alone does not establish whether the representation contains a structured temporal signal or merely separates the discrete gap classes observed by the probe.
We therefore train a numerical gap predictor only on
$\{0,20,40\}$ and evaluate it on unseen gaps
$\{10,30,50\}$.
The gaps 10 and 30 interpolate between values observed during probe training, while gap 50 additionally requires extrapolation beyond the largest training gap.

\begin{table}[!htbp]
\centering
\small
\begin{tabular}{ccc}
\toprule
Test gap & Endpoint difference & Joint target \\
\midrule
10 & \textbf{8.09} & 10.22 \\
30 & 10.93 & \textbf{4.39} \\
50 & 20.95 & \textbf{12.03} \\
\midrule
Overall & 13.32 & \textbf{8.88} \\
\bottomrule
\end{tabular}
\caption{Temporal-gap prediction on gap values absent during probe training. Values are mean absolute error in frames; lower is better.}
\label{tab:unseen_gap}
\end{table}

Overall, the joint target reduces MAE from 13.32 to 8.88 frames, a paired reduction of 4.44 frames with a 95\% confidence interval of $[3.87,5.00]$.
On the interpolation gaps $\{10,30\}$ alone, MAE decreases from 9.51 to 7.31 frames.
The joint target is not better at every individual gap, but its lower overall error indicates that the temporal signal is not limited to separating a fixed set of temporal classes and generalizes to unseen temporal separations.

\paragraph{Transition structure beyond endpoint displacement.}
The previous probes establish that the joint target contains a structured temporal relation, but they do not determine whether this information extends beyond the displacement between the two endpoints.
To isolate within-interval transition structure, we remove the straight-line component between the recorded robot states at the two endpoints.

For a transition with $\delta=50$, let
$s_t,\ldots,s_{t+50}$
denote the 14-D robot-state trajectory.
We define the straight-line interpolation
\begin{equation}
\bar{s}_{t+k}
=
\left(1-\frac{k}{50}\right)s_t
+
\frac{k}{50}s_{t+50},
\qquad
k=1,\ldots,49,
\end{equation}
and the residual trajectory
\begin{equation}
r_k
=
s_{t+k}-\bar{s}_{t+k}.
\end{equation}

The probe receives only the frozen visual representation of the two endpoints and predicts the complete $49\times14$ residual trajectory.
The prediction target therefore excludes the direct straight-line displacement from $s_t$ to $s_{t+50}$ and instead measures trajectory structure describing how the transition deviates from this endpoint change.
The encoder does not observe intermediate robot states, so the probe measures trajectory structure predictable from the relationship between the two visual endpoints rather than reconstruction of observed intermediate states.

\begin{table}[!htbp]
\centering
\small
\begin{tabular}{lccc}
\toprule
Target & Endpoint diff. & Joint target & Difference \\
\midrule
12-D arm residual & 0.488 & \textbf{0.581} & +0.093 \\
14-D state residual & 0.485 & \textbf{0.582} & +0.097 \\
\bottomrule
\end{tabular}
\caption{Decoding the trajectory residual after removing straight-line endpoint displacement. Higher mean $R^2$ is better.}
\label{tab:path_residual}
\end{table}

For the full 14-D trajectory, the joint target improves mean $R^2$ from 0.485 to 0.582.
The paired improvement is 0.097 with a 95\% confidence interval of $[0.082,0.112]$.
The joint target is also better at all 49 intermediate time steps, with per-step improvements ranging from 0.069 to 0.116.
Because the direct endpoint displacement has been removed from the prediction target, the result shows that the additional information in the joint representation is not limited to where the state starts and ends, but also reflects within-interval trajectory structure.

\paragraph{Endpoint-displacement control.}
As a complementary control, we directly predict the endpoint robot-state displacement,
\begin{equation}
\Delta s = s_{t+50}-s_t.
\end{equation}
Unlike the residual-trajectory target above, this objective is directly aligned with the subtraction used to construct the endpoint-difference representation.

\begin{table}[!htbp]
\centering
\small
\begin{tabular}{lcc}
\toprule
Representation & Mean $R^2$ $\uparrow$ & 95\% CI \\
\midrule
Endpoint difference & \textbf{0.740} & 0.711--0.777 \\
Joint target & 0.718 & 0.680--0.761 \\
\bottomrule
\end{tabular}
\caption{Direct endpoint-displacement prediction.}
\label{tab:endpoint_displacement}
\end{table}

Endpoint differencing is slightly better for this direct displacement target, with a paired Joint-minus-Difference effect of $-0.022$ and a 95\% confidence interval of $[-0.036,-0.009]$.
This reverse result clarifies the distinction between the two representations rather than suggesting that the joint target is uniformly superior.
Feature differencing provides a strong representation of first-order endpoint change, whereas joint encoding better preserves temporal relations and within-interval trajectory structure.

\paragraph{Summary.}
Together, the policy ablation and frozen-representation probes distinguish joint transition encoding from explicit endpoint subtraction.
Endpoint differencing performs substantially worse as policy supervision, with 70.9\% average success compared with 79.2\% for the joint target.
At the representation level, the joint target makes temporal separation more accessible, generalizes better to unseen temporal gaps, and better predicts trajectory structure after the endpoint displacement is removed.
These results indicate that the benefit of joint encoding is not explained by first-order endpoint difference alone.

\subsubsection{Spatial Diagnostic of the Joint Target}

The preceding experiments analyze what temporal information is accessible from the joint target.
We additionally examine whether the patch-wise representation changes induced by the future observation are spatially related to regions that change in the scene.
This analysis is intended as a qualitative and complementary diagnostic rather than a motion-segmentation objective.

A direct comparison between a two-frame joint representation and independently encoded single-image representations can confound temporal interaction with image--video encoding differences.
We therefore use a matched static control in the same two-frame encoding mode.
For a transition $(O_t,O_{t+50})$, we compare the dynamic joint target
\begin{equation}
Y_{\mathrm{joint}}
=
E_J
\left(
\operatorname{Stack}_{\mathrm{time}}
(O_t,O_{t+50})
\right)
\end{equation}
with a static-current target
\begin{equation}
Y_{\mathrm{static}}
=
E_J
\left(
\operatorname{Stack}_{\mathrm{time}}
(O_t,O_t)
\right).
\end{equation}

For each spatial patch $p$, we define the representation change as
\begin{equation}
r_p
=
1-
\cos
\left(
Y_{\mathrm{joint},p},
Y_{\mathrm{static},p}
\right).
\end{equation}
We compare this patch-wise residual with the absolute RGB change between the current and future observations, downsampled to the same spatial patch grid.

The diagnostic uses 200 held-out test episodes with five transitions per episode, for 1,000 transitions in total.
Table~\ref{tab:spatial_diagnostic} reports the median patch-wise Pearson correlation with RGB change.

\begin{table}[!htbp]
\centering
\small
\begin{tabular}{lc}
\toprule
Spatial comparison & Median correlation \\
\midrule
Joint vs. matched static current & \textbf{0.279} \\
Joint vs. matched static future & 0.182 \\
Joint vs. matched static endpoint mean & 0.190 \\
\bottomrule
\end{tabular}
\caption{Patch-wise correlation between representation change and RGB change under matched static controls.}
\label{tab:spatial_diagnostic}
\end{table}

\begin{figure}[!htbp]
\centering
\includegraphics[width=0.7\linewidth]{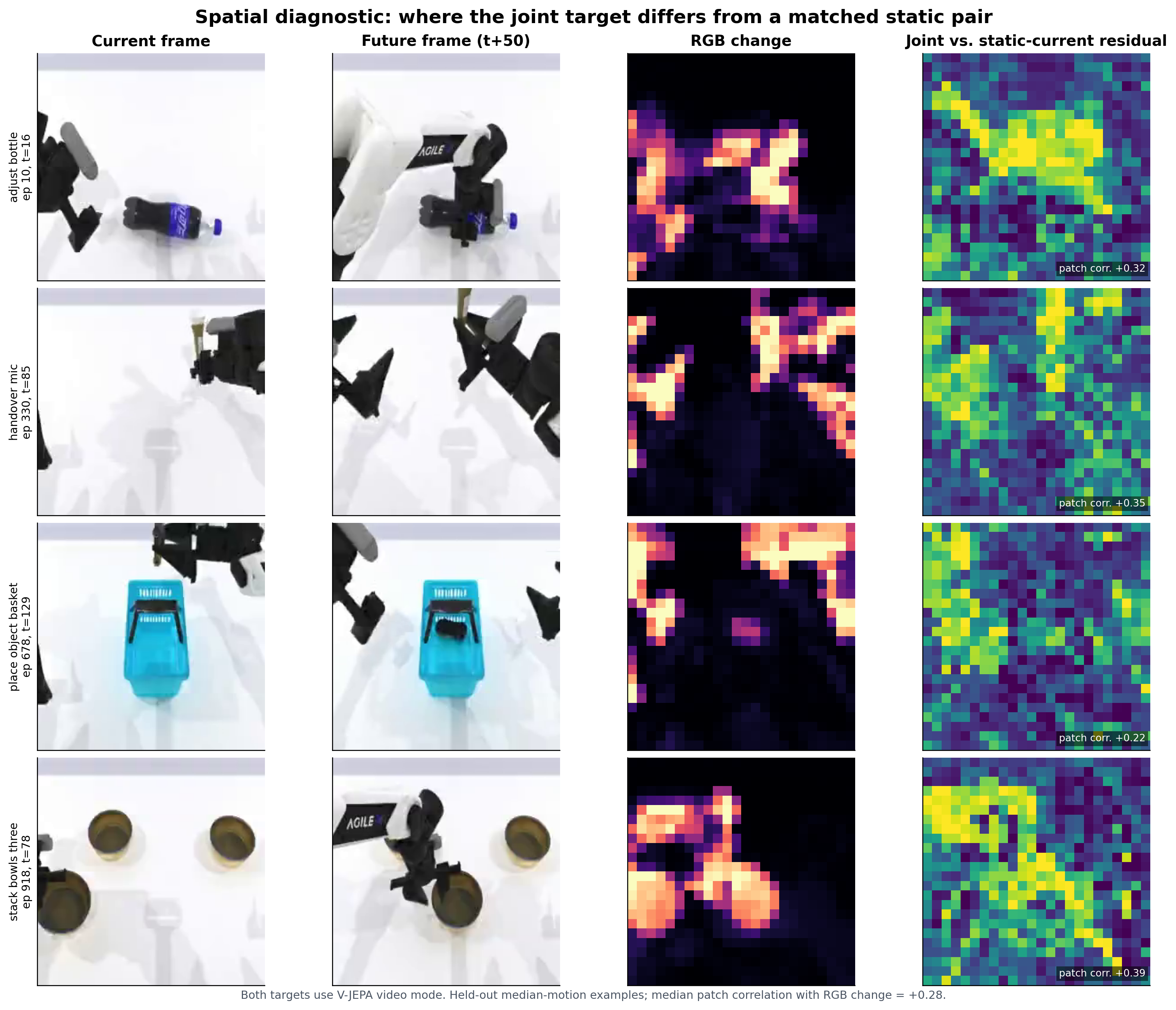}
\caption{
Spatial diagnostic of the joint target.
Patch-wise representation changes are compared with image-space changes between the current and future observations.
The residual tends to be stronger around visually changing regions, while remaining more spatially distributed than a pixel-level motion map.
}
\label{fig:spatial_diagnostic}
\end{figure}

The positive correlation indicates that part of the patch-wise change in the joint representation is spatially associated with regions undergoing visual change.
However, the correspondence is moderate rather than exact.
This is expected because the target represents high-level visual structure and may capture changes in object configuration, occlusion, contact, and contextual relationships beyond local RGB displacement.
We therefore interpret this analysis as supporting spatially structured transition information, rather than as evidence of precise motion localization.

\FloatBarrier
\subsection{Inference Efficiency}

Table~\ref{tab:inference_efficiency} reports recorded inference latency under the same RoboTwin inference setting. 
At deployment, JEPA-WAM does not execute target encoding, latent transition prediction, or iterative future-frame generation. 
The target branch and transition loss are used only during training.

\begin{table}[!htbp]
\centering
\small
\begin{tabular}{lcc}
\toprule
Method & Median Latency (ms) $\downarrow$ & Frequency (Hz) $\uparrow$ \\
\midrule
$\pi_{0.5}$ & 54.05 & 18.50 \\
$\pi_{0.5}$ + JEPA Obj. & 55.12 & 18.14 \\
JEPA-WAM (Ours) & 85.00 & 11.76 \\
ABot-M0 & 125.23 & 7.99 \\
\bottomrule
\end{tabular}
\caption{Inference latency and frequency in the RoboTwin setting.}
\label{tab:inference_efficiency}
\end{table}
\clearpage
\section{Complete RoboTwin 2.0 Results}
\label{app:robotwin-full}

Complete task-wise success rates for all 20 RoboTwin 2.0 tasks are reported below.
Methods are grouped according to whether they use large-scale robot-policy pretraining.

\begin{table}[!htbp]
\centering
\setlength{\tabcolsep}{4pt}
\renewcommand{\arraystretch}{1.15}
\begin{tabular}{@{}lcccccccc@{}}
\toprule
& \multicolumn{2}{c}{ACT}
& \multicolumn{2}{c}{DP}
& \multicolumn{2}{c}{DP3}
& \multicolumn{2}{c}{JEPA-WAM} \\
Task
& Cle. & Ran.
& Cle. & Ran.
& Cle. & Ran.
& Cle. & Ran. \\
\midrule

Adjust Bottle
& 97 & 23
& 97 & 0
& \textbf{99} & 3
& \textbf{99} & \textbf{87} \\

Beat Block Hammer
& 56 & 3
& 42 & 0
& \textbf{72} & 8
& 66 & \textbf{13} \\

Click Alarmclock
& 32 & 4
& 61 & 5
& 77 & 14
& \textbf{97} & \textbf{45} \\

Click Bell
& 58 & 3
& 54 & 0
& 90 & 0
& \textbf{100} & \textbf{39} \\

Dump Bin Bigbin
& 68 & 1
& 49 & 0
& 85 & 53
& \textbf{94} & \textbf{63} \\

Grab Roller
& 94 & 25
& \textbf{98} & 0
& \textbf{98} & 2
& 92 & \textbf{79} \\

Handover Mic
& 85 & 0
& 53 & 0
& \textbf{100} & 3
& 98 & \textbf{14} \\

Lift Pot
& 88 & 0
& 39 & 0
& \textbf{97} & 0
& 91 & \textbf{30} \\

Place Bread Basket
& 6 & 0
& 14 & 0
& 26 & 1
& \textbf{38} & \textbf{6} \\

Place Bread Skillet
& 7 & 0
& 11 & 0
& 19 & 0
& \textbf{42} & \textbf{2} \\

Place Burger Fries
& 49 & 0
& 72 & 0
& 72 & \textbf{18}
& \textbf{98} & 14 \\

Place Cans Plasticbox
& 16 & 0
& 40 & 0
& 48 & 3
& \textbf{64} & \textbf{20} \\

Place Empty Cup
& 61 & 0
& 37 & 0
& 65 & 1
& \textbf{92} & \textbf{8} \\

Place Object Basket
& 15 & 0
& 15 & 0
& \textbf{65} & 0
& 62 & \textbf{14} \\

Place Shoe
& 5 & 0
& 23 & 0
& \textbf{58} & 2
& 36 & \textbf{10} \\

Press Stapler
& 31 & 6
& 6 & 0
& 69 & 3
& \textbf{88} & \textbf{70} \\

Shake Bottle Horizontally
& 63 & 4
& 59 & 18
& \textbf{100} & 25
& 94 & \textbf{55} \\

Shake Bottle
& 74 & 10
& 65 & 8
& \textbf{98} & 19
& 94 & \textbf{55} \\

Stack Bowls Three
& 48 & 0
& \textbf{63} & 0
& 57 & 5
& 59 & \textbf{37} \\

Stack Bowls Two
& 82 & 0
& 61 & 0
& 83 & 6
& \textbf{94} & \textbf{76} \\

\midrule
Average
& 51.8 & 4.0
& 48.0 & 1.6
& 73.9 & 8.3
& \textbf{79.9} & \textbf{36.9} \\
\bottomrule
\end{tabular}

\caption{
Complete task-wise success rates (\%) on RoboTwin 2.0 for methods without large-scale robot-policy pretraining.
Cle. and Ran. denote the Clean and Random settings, respectively.
The best result for each task and setting is shown in bold.
}
\label{tab:robotwin-full-nopt}
\end{table}

\begin{table}[!htbp]
\centering
\setlength{\tabcolsep}{4pt}
\renewcommand{\arraystretch}{1.15}
\begin{tabular}{@{}lcccccccc@{}}
\toprule
& \multicolumn{2}{c}{RDT-1B}
& \multicolumn{2}{c}{$\pi_0$}
& \multicolumn{2}{c}{$\pi_{0.5}$}
& \multicolumn{2}{c}{$\pi_{0.5}$ + JEPA Obj.} \\
Task
& Cle. & Ran.
& Cle. & Ran.
& Cle. & Ran.
& Cle. & Ran. \\
\midrule

Adjust Bottle
& 81 & \textbf{75}
& 90 & 56
& 98 & 26
& \textbf{100} & 31 \\

Beat Block Hammer
& \textbf{77} & \textbf{37}
& 43 & 21
& 76 & 3
& 69 & 5 \\

Click Alarmclock
& 61 & 12
& 63 & 11
& 90 & \textbf{63}
& \textbf{94} & 43 \\

Click Bell
& 80 & 9
& 44 & 3
& 97 & \textbf{58}
& \textbf{100} & 48 \\

Dump Bin Bigbin
& 64 & 32
& 83 & 24
& \textbf{95} & 41
& 90 & \textbf{58} \\

Grab Roller
& 74 & 43
& 96 & \textbf{80}
& 99 & 64
& \textbf{100} & 54 \\

Handover Mic
& 90 & \textbf{31}
& \textbf{98} & 13
& 84 & 8
& 91 & 17 \\

Lift Pot
& 72 & 9
& \textbf{84} & \textbf{36}
& 63 & 4
& 80 & 4 \\

Place Bread Basket
& 10 & 2
& 17 & 4
& 51 & \textbf{30}
& \textbf{77} & 29 \\

Place Bread Skillet
& 5 & 1
& 23 & 1
& 56 & 20
& \textbf{59} & \textbf{21} \\

Place Burger Fries
& 50 & 27
& 80 & 4
& 83 & 54
& \textbf{96} & \textbf{59} \\

Place Cans Plasticbox
& 6 & 5
& 34 & 2
& 36 & \textbf{42}
& \textbf{95} & 40 \\

Place Empty Cup
& 56 & 7
& 37 & 11
& 77 & \textbf{59}
& \textbf{100} & 54 \\

Place Object Basket
& 33 & \textbf{17}
& 16 & 2
& 66 & 3
& \textbf{73} & 6 \\

Place Shoe
& \textbf{35} & 7
& 28 & 6
& 26 & \textbf{15}
& 34 & 13 \\

Press Stapler
& 41 & 24
& 62 & \textbf{29}
& 66 & 22
& \textbf{86} & 22 \\

Shake Bottle Horizontally
& 84 & 51
& 99 & 51
& \textbf{100} & 82
& \textbf{100} & \textbf{86} \\

Shake Bottle
& 74 & 45
& 97 & 60
& 99 & 82
& \textbf{100} & \textbf{84} \\

Stack Bowls Three
& 51 & 17
& \textbf{66} & 24
& 59 & 29
& 55 & \textbf{30} \\

Stack Bowls Two
& 76 & 30
& 91 & 41
& 87 & 40
& \textbf{93} & \textbf{45} \\

\midrule
Average
& 56.0 & 24.1
& 62.5 & 23.9
& 75.4 & 37.2
& \textbf{84.6} & \textbf{37.5} \\
\bottomrule
\end{tabular}
\caption{
Complete task-wise success rates (\%) on RoboTwin 2.0 for methods with large-scale robot-policy pretraining.
Cle. and Ran. denote the Clean and Random settings, respectively.
The best result for each task and setting is shown in bold.
}
\label{tab:robotwin-full-pt}
\end{table}
\clearpage
\section{Real-World Detailed Results}
\label{app:real-world-results}

\subsection{Task-Level Summary}

Table~\ref{tab:real-world-summary} reports the task-wise normalized completion scores for all methods under both ID and OOD settings.
Each value is averaged over 10 real-world rollouts, and the average is computed across the five tasks.
The complete rollout-level records are provided in the following subsection.

\begin{table}[!htbp]
\centering
\setlength{\tabcolsep}{5pt}
\renewcommand{\arraystretch}{1.08}
\begin{tabular}{@{}llrrrrrr@{}}
\toprule
Method & Setting & Bread & Fruit & Place-3 & Stack-3 & Drawer & Avg. \\
\midrule
$\pi_0$
& ID
& 90.00 & 52.50 & 36.60 & 35.00 & 45.00 & 51.82 \\
$\pi_0$
& OOD
& 20.00 & 20.00 & 40.00 & 10.00 & 22.50 & 22.50 \\
\midrule
JEPA-WAM
& ID
& 90.00 & 60.00 & 46.60 & 55.00 & 47.50 & 59.82 \\
JEPA-WAM
& OOD
& 80.00 & 55.00 & 53.40 & 40.00 & 42.50 & 54.18 \\
\midrule
$\pi_{0.5}$
& ID
& 100.00 & 70.00 & 70.10 & 70.00 & 77.50 & 77.52 \\
$\pi_{0.5}$
& OOD
& 90.00 & 65.00 & 70.00 & 65.00 & 72.50 & 72.50 \\
\midrule
$\pi_{0.5}$+JEPA Obj.
& ID
& 100.00 & 85.00 & 96.70 & 75.00 & 95.00 & 90.34 \\
$\pi_{0.5}$+JEPA Obj.
& OOD
& 100.00 & 70.00 & 93.40 & 75.00 & 85.00 & 84.68 \\
\bottomrule
\end{tabular}
\caption{
Mean normalized task-completion scores (\%) over 10 real-world rollouts per task and setting.
}
\label{tab:real-world-summary}
\end{table}

\subsection{Per-Rollout Records}

Tables~\ref{tab:real-pi0-records}--\ref{tab:real-pi05-jepa-records} report the complete rollout-level records for all four policies.
Br., Fr., Pl., St., and Dr. denote bread placement, fruit placement, three-block placement, three-block stacking, and drawer manipulation, respectively.
All scores are normalized to $[0,1]$, and the final row reports the task-wise mean over the 10 rollouts.

\begin{table}[!htbp]
\centering
\setlength{\tabcolsep}{4pt}
\renewcommand{\arraystretch}{1.08}
\begin{tabular}{@{}rcccccccccc@{}}
\toprule
& \multicolumn{5}{c}{ID}
& \multicolumn{5}{c}{OOD} \\
\cmidrule(lr){2-6}
\cmidrule(lr){7-11}
Run
& Br. & Fr. & Pl. & St. & Dr.
& Br. & Fr. & Pl. & St. & Dr. \\
\midrule
1  & 1 & 0    & 1    & 0    & 1    & 1 & 0    & 1    & 0    & 0    \\
2  & 1 & 0.75 & 0    & 0    & 1    & 0 & 0    & 1    & 0    & 0.25 \\
3  & 1 & 1    & 0.33 & 1    & 0    & 0 & 0.50 & 0    & 0    & 0    \\
4  & 1 & 0    & 1    & 0.50 & 0    & 0 & 0    & 0.33 & 0.50 & 0    \\
5  & 0 & 0.50 & 0    & 0.50 & 0    & 0 & 0    & 0.67 & 0    & 1    \\
6  & 1 & 0.50 & 0    & 0    & 0.50 & 0 & 0.50 & 0    & 0    & 0    \\
7  & 1 & 0.25 & 0.67 & 0.50 & 0.25 & 0 & 0.50 & 0.33 & 0 & 0    \\
8  & 1 & 0.75 & 0    & 1    & 0    & 1 & 0    & 0    & 0    & 0.25 \\
9  & 1 & 1    & 0.33 & 0    & 1    & 0 & 0    & 0    & 0.50 & 0.50 \\
10 & 1 & 0.50 & 0.33 & 0    & 0.75 & 0 & 0.50 & 0.67 & 0 & 0.25 \\
\midrule
Avg.
& 0.90 & 0.525 & 0.366 & 0.35 & 0.45
& 0.20 & 0.20 & 0.40 & 0.10 & 0.225 \\
\bottomrule
\end{tabular}
\caption{Per-rollout real-world results for $\pi_0$.}
\label{tab:real-pi0-records}
\end{table}

\begin{table}[!htbp]
\centering
\setlength{\tabcolsep}{4pt}
\renewcommand{\arraystretch}{1.08}
\begin{tabular}{@{}rcccccccccc@{}}
\toprule
& \multicolumn{5}{c}{ID}
& \multicolumn{5}{c}{OOD} \\
\cmidrule(lr){2-6}
\cmidrule(lr){7-11}
Run
& Br. & Fr. & Pl. & St. & Dr.
& Br. & Fr. & Pl. & St. & Dr. \\
\midrule
1  & 1 & 0.50 & 0.33 & 0    & 0.25 & 1 & 0.50 & 0    & 0    & 0.50 \\
2  & 1 & 0.50 & 0.67 & 0.50 & 0.50 & 1 & 0.50 & 1    & 0    & 0.25 \\
3  & 0 & 1    & 0    & 1    & 0.25 & 1 & 0.50 & 0    & 0.50 & 0.50 \\
4  & 1 & 0    & 1    & 0    & 0    & 0 & 1    & 0.33 & 0    & 1    \\
5  & 1 & 0.50 & 1    & 1    & 0.25 & 1 & 0    & 0.67 & 1    & 1    \\
6  & 1 & 1    & 0.67 & 1    & 1    & 1 & 1    & 0    & 0.50 & 0    \\
7  & 1 & 1    & 0    & 0.50 & 0    & 0 & 1    & 0.67 & 0    & 0.25 \\
8  & 1 & 1    & 0.33 & 0    & 1    & 1 & 0.50 & 0.67 & 0.50 & 0    \\
9  & 1 & 0.50 & 0.33 & 1    & 0.50 & 1 & 0    & 1    & 1    & 0.25 \\
10 & 1 & 0    & 0.33 & 0.50 & 1    & 1 & 0.50 & 1    & 0.50 & 0.50 \\
\midrule
Avg.
& 0.90 & 0.60 & 0.466 & 0.55 & 0.475
& 0.80 & 0.55 & 0.534 & 0.40 & 0.425 \\
\bottomrule
\end{tabular}
\caption{Per-rollout real-world results for JEPA-WAM.}
\label{tab:real-jepa-records}
\end{table}

\begin{table}[!htbp]
\centering
\setlength{\tabcolsep}{4pt}
\renewcommand{\arraystretch}{1.08}
\begin{tabular}{@{}rcccccccccc@{}}
\toprule
& \multicolumn{5}{c}{ID}
& \multicolumn{5}{c}{OOD} \\
\cmidrule(lr){2-6}
\cmidrule(lr){7-11}
Run
& Br. & Fr. & Pl. & St. & Dr.
& Br. & Fr. & Pl. & St. & Dr. \\
\midrule
1  & 1 & 1    & 1    & 1    & 1    & 1 & 1    & 1    & 0    & 1    \\
2  & 1 & 0.50 & 0.67 & 1    & 1    & 1 & 0    & 1    & 1    & 1    \\
3  & 1 & 0    & 0.33 & 0.50 & 0.50 & 0 & 1    & 0.67 & 1    & 0.75 \\
4  & 1 & 1    & 1    & 1    & 1    & 1 & 1    & 0.33 & 0    & 0.50 \\
5  & 1 & 0.50 & 0.33 & 0    & 0.25 & 1 & 0.50 & 0.67 & 0.50 & 1    \\
6  & 1 & 1    & 1    & 1    & 0.50 & 1 & 0    & 0.33 & 1    & 0.25 \\
7  & 1 & 0    & 0.67 & 1    & 1    & 1 & 1    & 0.67 & 0.50 & 0.50 \\
8  & 1 & 1    & 0.67 & 1    & 1    & 1 & 1    & 1    & 0.50 & 1    \\
9  & 1 & 1    & 0.67 & 0    & 1    & 1 & 0.50 & 0.33 & 1    & 1    \\
10 & 1 & 1    & 0.67 & 0.50 & 0.50 & 1 & 0.50 & 1    & 1    & 0.25 \\
\midrule
Avg.
& 1.00 & 0.70 & 0.701 & 0.70 & 0.775
& 0.90 & 0.65 & 0.70 & 0.65 & 0.725 \\
\bottomrule
\end{tabular}
\caption{Per-rollout real-world results for $\pi_{0.5}$.}
\label{tab:real-pi05-records}
\end{table}

\begin{table}[!htbp]
\centering
\setlength{\tabcolsep}{4pt}
\renewcommand{\arraystretch}{1.08}
\begin{tabular}{@{}rcccccccccc@{}}
\toprule
& \multicolumn{5}{c}{ID}
& \multicolumn{5}{c}{OOD} \\
\cmidrule(lr){2-6}
\cmidrule(lr){7-11}
Run
& Br. & Fr. & Pl. & St. & Dr.
& Br. & Fr. & Pl. & St. & Dr. \\
\midrule
1  & 1 & 1    & 1    & 1    & 1    & 1 & 0    & 1    & 1    & 1    \\
2  & 1 & 1    & 1    & 0.50 & 1    & 1 & 0.50 & 1    & 1    & 1    \\
3  & 1 & 0.50 & 1    & 1    & 1    & 1 & 1    & 1    & 0    & 1    \\
4  & 1 & 1    & 1    & 1    & 1    & 1 & 0.50 & 1    & 0.50 & 1    \\
5  & 1 & 1    & 1    & 0.50 & 1    & 1 & 1    & 0.67 & 1    & 0    \\
6  & 1 & 1    & 1    & 0.50 & 0.75 & 1 & 1    & 1    & 1    & 0.50 \\
7  & 1 & 0.50 & 0.67 & 1    & 1    & 1 & 0.50 & 0.67 & 1    & 1    \\
8  & 1 & 0.50 & 1    & 1    & 1    & 1 & 1    & 1    & 0.50 & 1    \\
9  & 1 & 1    & 1    & 0.50 & 1    & 1 & 1    & 1    & 1    & 1    \\
10 & 1 & 1    & 1    & 0.50 & 0.75 & 1 & 0.50 & 1    & 0.50 & 1    \\
\midrule
Avg.
& 1.00 & 0.85 & 0.967 & 0.75 & 0.95
& 1.00 & 0.70 & 0.934 & 0.75 & 0.85 \\
\bottomrule
\end{tabular}
\caption{Per-rollout real-world results for $\pi_{0.5}$ + JEPA Obj..}
\label{tab:real-pi05-jepa-records}
\end{table}


\end{document}